\documentclass{article}

\usepackage{arxiv}

\usepackage[utf8]{inputenc} 
\usepackage[T1]{fontenc}    
\usepackage{hyperref}       
\usepackage{url}            
\usepackage{booktabs}       
\usepackage{amsfonts,amsmath,amssymb}       
\usepackage{nicefrac}       
\usepackage{microtype}      
\usepackage{graphicx}
\usepackage{caption}

\usepackage{color, colortbl}

\usepackage{mathptmx,amsthm}
\usepackage{xcolor}
\usepackage{multirow}
\usepackage{subcaption}
\usepackage{tabularx,adjustbox}

\newcommand{\ie}{\textit{i.e.,}}
\newcommand{\eg}{\textit{e.g.,}}
\newcommand{\commenttxt}[1]{}

\newcommand{\mybar}{\kern1pt\rule[-\dp\strutbox]{.8pt}{\baselineskip}\kern1pt}
\newcommand{\citep}{\cite}

\renewcommand{\mytagline}{} 

\theoremstyle{definition}
\newtheorem{definition}{Definition}[section]

\title{Fusion and Grouping Strategies in Deep Learning for Local Climate Zone Classification of Multimodal Remote Sensing Data}
\author{
  Ancymol Thomas\textsuperscript{1},
  Jaya~Sreevalsan-Nair\textsuperscript{1}\thanks{\texttt{jnair@iiitb.ac.in}} 
  \\
  \textsuperscript{1}Graphics-Visualization-Computing Lab,\\
  International Institute of Information Technology Bangalore, Karnataka 560100, India. \\
  \texttt{http://www.iiitb.ac.in/gvcl} \\
}

\begin{document}
\maketitle

\begin{abstract}
Local Climate Zones (LCZs) give a zoning map to study urban structures and land use and analyze the impact of urbanization on local climate. Multimodal remote sensing enables LCZ classification, for which data fusion is significant for improving accuracy owing to the data complexity. However, there is a gap in a comprehensive analysis of the fusion mechanisms used in their deep learning (DL) classifier architectures. This study analyzes different fusion strategies in the multi-class LCZ classification models for multimodal data and grouping strategies based on inherent data characteristics. The different models involving Convolutional Neural Networks (CNNs) include: (i) baseline hybrid fusion (FM1), (ii) with self- and cross-attention mechanisms (FM2), (iii) with the multi-scale Gaussian filtered images (FM3), and (iv) weighted decision-level fusion (FM4). Ablation experiments are conducted to study the pixel-, feature-, and decision-level fusion effects in the model performance. Grouping strategies include band grouping (BG) within the data modalities and label merging (LM) in the ground truth. Our analysis is exclusively done on the So2Sat LCZ42 dataset, which consists of Synthetic Aperture Radar (SAR) and Multispectral Imaging (MSI) image pairs. Our results show that FM1 consistently outperforms simple fusion methods. FM1 with BG and LM is found to be the most effective approach among all fusion strategies, giving an overall accuracy of 76.6\%. Importantly, our study highlights the effect of these strategies in improving prediction accuracy for the underrepresented classes. Our code and processed datasets are available at \url{https://github.com/GVCL/LCZC-MultiModalHybridFusion}
\end{abstract}


\section{Introduction} \label{sec:introduction}
\begin{figure}[ht]
\centering
\includegraphics[width=\columnwidth]{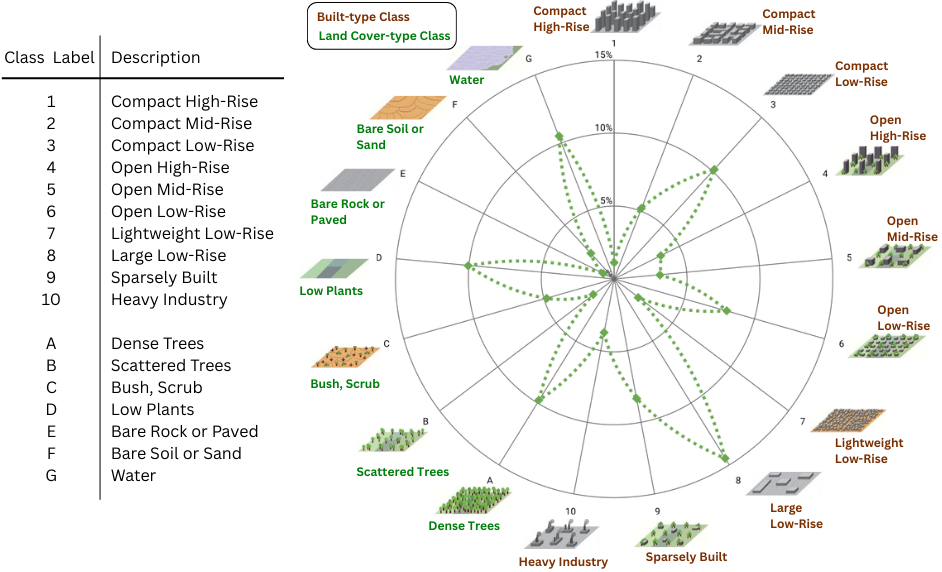}
\captionof{figure}{The 17-class list of LCZs and an illustrated class distribution radar plot visualization of the So2Sat LCZ42 dataset~\citep{zhu2020so2sat}.}
\label{fig:dataset}
\end{figure}

The adverse effects of rapid urbanization include environmental pollution and increasingly uninhabitable conditions, \eg~Urban Heat Islands. This leads to a vicious cycle affecting urban micro- and macro-climate, urban environmental quality, and human health. Local climate zones (LCZs) form a zoning map that segregates and classifies space based on levels of urbanization, urban structures, and land-use patterns in a global standard framework~\citep{stewart2012local,bechtel2015mapping}. The seventeen LCZ classes include land-cover type (urban, vegetation, water), their structures (height and density of buildings), and anthropogenic thermal emissions~\citep{stewart2012local,bechtel2015mapping,qiu2019local}. The So2Sat LCZ42 dataset~\citep{zhu2020so2sat} is one of the earliest publicly available benchmark multimodal remote sensing datasets for the 17-class LCZ classification task (Figure~\ref{fig:dataset}).

Amongst several remote sensing technologies, Synthetic Aperture Radar (SAR) and Multispectral Imaging (MSI) are complementary and improve information extraction from the data, \eg~for LCZ classification task~\citep{zhu2020so2sat}. SAR and MSI use \textit{active} and \textit{passive} sensing, respectively~\citep{lillesand2015remote}. SAR emits microwaves of varying wavelengths (depending on C-, L-, or X-band) and penetrates through \textit{cloud cover} and \textit{participating media}, thus, providing all-day, all-weather availability; and penetrates, to a lesser extent, through the Earth's surface to get the subsurface features. MSI is an optical imaging technology that estimates the energy reflected and radiated by the Earth's surface at different wavelengths, \ie~\textit{spectral bands}. SAR images provide coherent information on the moisture, roughness, and terrain structure of the observed surface, and hence, are widely used for land-use/land-cover (LULC) mapping. However, unlike optical images, SAR images are challenging to interpret due to limited band numbers and the effects caused by speckle noise, slant-range imaging, foreshortening, layover, and shadows. The higher spectral resolution in MSI provides information on biological and chemical properties of scanned regions, \eg~vegetation analysis and surface water segmentation. Overall, SAR-MSI image pairs maximize the information retrieval on physical and spectral properties of urban regions at all times, needed for LULC and LCZ classifications~\citep{zhu2020so2sat}.  

Multi-sensor data, treated as \textit{multimodal data}, is used as an input to DL models for LCZ classification of their corresponding regions~\citep{zhu2020so2sat,qiu2020multilevel}. However, the dataset requires sophisticated learning models owing to its complexity in the multimodal data arising from the differences in imaging mechanisms of multiple sources. To alleviate the challenges in data complexity, fusion DL models are increasingly seen as a viable solution for LCZ classification of multimodal remote sensing data~\citep{he2023sar}. Upon fusion, the dataset is enhanced to provide improved information of the Earth surface and structural characteristics~\citep{kulkarni2020pixel} and classification outcomes~\citep{faqeimproving2023,gaetano2017fusion}. For multimodal image datasets, the fusion is performed at three levels: \textit{pixel}, \textit{feature}, and \textit{decision}. Pixel-level techniques directly integrate the information from input images~\citep{yang2007overview}. Feature-level techniques involve the extraction of relevant features such as pixel intensities, textures, or edges, that are combined to create additional merged features~\citep{duan2024deep}. 

\begin{figure}[t]
\centering
\includegraphics[width=\columnwidth]{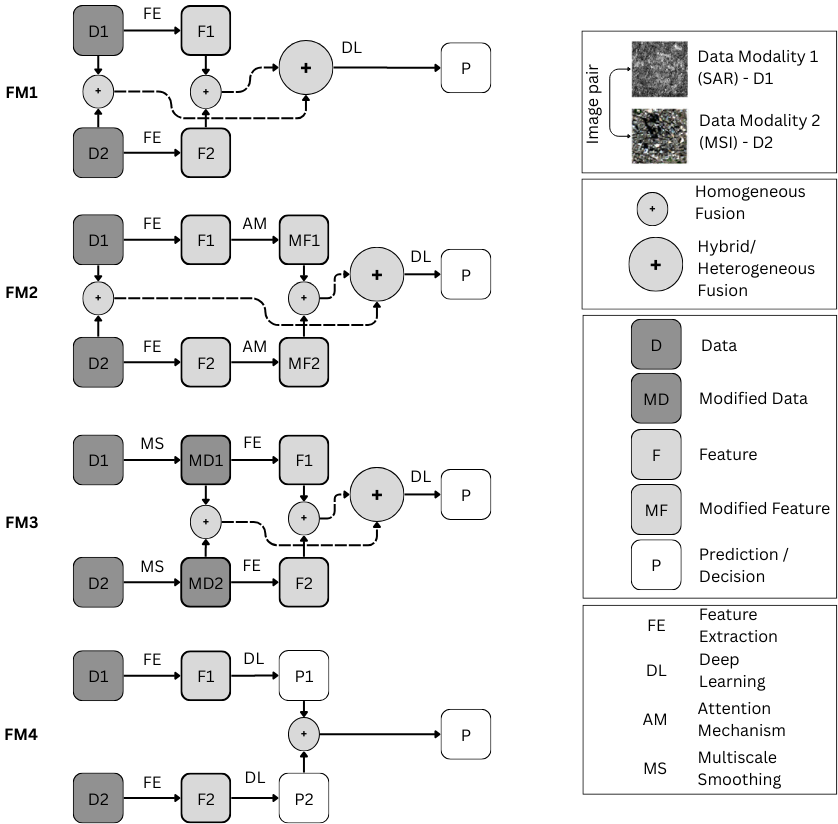}
\caption{A summary of the proposed hybrid fusion models, where F1 is the baseline, F2 is with an attention mechanism, F3 is with multi-scale Gaussian smoothing, and F3 is a late-fusion model.}
\label{fig:fusionmodels}
\end{figure}

Convolutional Neural Networks (CNNs) are apt for LCZ classification as they use multiple convolutional kernel sizes to extract spatial-spectral information from the input data. However, CNNs fall short in handling high-dimensional data and sequential characteristics in spectral data, leading to inefficiencies in contextual information extraction~\citep{dang2023double}. As one of the solutions, self- and cross-attention mechanisms enable learning cross-feature dependencies in multi-source remote sensing data~\citep{li2024crossfuse, hussain2025cross}. Another solution is a multi-scale approach where multi-scale feature extraction provides varied information at different spatial scales, which, upon integration, provides a comprehensive representation of image data with a spatial context~\citep{huang2021multiscale}. 

Despite the advances in fusion models for LCZ classification, these models are not assessed to analyze the effect of specific fusion or grouping strategies on class-wise accuracies. In addition to high imaging-based variations in different modalities of the dataset, there is a significant class imbalance in real-world datasets, which poses challenges in model performance. For instance, in the So2Sat LCZ42 dataset, classes 1, 4, 5, 7, B, E, and F have less than 2\% samples, whereas class 8 has $\sim$15\% (Figure~\ref{fig:dataset}). Improving class-wise accuracy for underrepresented and under-performing classes is essential for improving overall accuracy. Hence, we address the gap in fine-grained analysis of fusion and grouping strategies in class-wise prediction outcomes.

 In this work, we propose a comparative study of different fusion strategies for LCZ classification using SAR-MSI image pairs in the So2Sat LCZ42 dataset. Building on the baseline \textit{hybrid fusion}, \ie~data- and feature-level fusion, model (FM1), two enhancements are introduced: (a) integration of attention mechanisms (FM2), and (b) application of multi-scale Gaussian filtering (FM3). We compare these models with a simple \textit{decision-level fusion} strategy using a weighted U-Net-CNN architecture (FM4). A summary of the pipelines for the proposed fusion models is given in Figure~\ref{fig:fusionmodels}. Ablation experiments are conducted to examine the effects of pixel-, feature-, and decision-level fusion. In addition to fusion strategies in the model architectures, we propose data- and label-level grouping strategies in the inputs and ground truth, respectively. At the data-level, we perform spectral band grouping (SBG) to reduce the misclassification arising from spectrally similar bands in the data. Similarly, at the label-level, we perform label merging (LM) to reduce the misclassification arising from semantically similar classes in the ground truth.
 
The key contributions of this study are: 
\begin{itemize}
  \item To compare different fusion strategies of SAR and MSI data from Sentinel-1 (S1) and Sentinel-2 (S2), respectively, for LCZ classification, including attention mechanism, multiscle Gaussian smoothing, and late fusion,
  \item To evaluate the effect of spectral band grouping (SBG) on LCZ classification,
  \item To evaluate the effect of label merging (LM) on LCZ classification.
\end{itemize}

\section{Related Work} \label{sec:related_work}
Here, we consider the state of the art in LCZ classification, multimodal data fusion, attention-based models, and multi-scale Gaussian smoothing.

\subsection{LCZ Classification}
LCZ classification task has been implemented on different remote sensing datasets, using DL models as widely preferred tools. Biclassified CNN variants \textemdash~MobileNet-SegNet (MS), MobileNet-U-Net (MU), and MobileNet-PSPNet (MP) \textemdash~are evaluated for LCZ classification using very-high-resolution (VHR) images~\citep{liu2023comparison}, of which MP was the most accurate model, and MU was the most optimal model. While there is work on datasets with single modality, multimodal data is found to provide better outcomes for LCZ classification. 

There have been multimodal LCZ studies with multispectral images from S2 or Landsat 8 and SAR data. LCZ classification using CNN for the fusion of SAR data and optical images showed improved accuracy~\citep{hu2018feature}. There is work on decision-level fusion strategies such as the Dual-stream Fusion framework (DF4LCZ)~\citep{wu2024df4lcz}, which integrates Google imagery with spatial–spectral features from S2 images for LCZ classification. MsF-LCZ-Net model~\citep{he2023sar} is a recent CNN-based network for multimodal feature extraction and fusion, which is followed by a classifier for LCZ prediction. This model increased the overall accuracy by 2\%, but the class-wise accuracy for a few of the classes, specifically LCZ C ("Bush, scrub"), is significantly less (3.7\%). This work also explored a band grouping method where SAR image bands are grouped into polarization-based groups (VV, VH, and CMOE), while MS image bands are grouped into four spectral ranges \textemdash~RGB, VRE, SWIR, and NIR \textemdash~based on band correlations, enabling more effective multi-source feature fusion. This spectral band grouping strategy is adopted in our systematic analysis of fusion models.

CNN-based data fusion is underexplored for LCZ classification. In the available studies, while the overall accuracy improves, the class-wise performance declines. This differential outcome points to the challenge of distinguishing between urban and natural classes. Our work bridges the gap in class-wise performance analysis. We also adopt fusion strategies, such as weighted fusion approaches that combine CNNs with graph attention networks for hyperspectral image classification~\citep{dong2022weighted}.

\subsection{Multimodal Data Fusion}
There have been studies on the fusion of microwave and optical sensor data in applications such as land cover and urban area mapping. Advanced machine learning techniques like SVM, ensemble learning, and deep learning enhance spatial data analysis and land cover classification~\citep{du2020advances}. A PCA-based image fusion technique yielded an improved composite image~\citep{kaur2021image}. Image fusion of SAR and MSI data using non-subsampled shearlet transform and activity measure improved the interpretability of SAR imagery~\citep{huang2022image}, to use them in scenarios where easily interpretable optical imagery is difficult to obtain.

Classification of multitemporal S1 and S2 imagery is improved using Extreme Gradient Boosting for Land Cover mapping~\citep{dobrinic2020integration}. Random Forest classifier is effective for pixel-based land cover classification of S1 and S2  data~\citep{castro2017joint}. DL models are extensively used within the remote sensing community for image fusion, \eg~the first hyperspectral-SAR data fusion DL model~\citep{hu2017fusionet} used a two-branch architecture that separately extracts heterogeneous features for final convolutional fusion.

\subsection{Attention-based Models}
The transformer model is widely used currently for its ability to model contextual dependencies globally across different positions in a sequence. Transformers have been successfully adapted to computer vision, including remote sensing image analysis. To enhance spectral feature extraction, models such as SpectralFormer~\citep{hong2021spectralformer} have been introduced, using self-attention mechanisms to improve the modeling of spectral sequences. CrossT-Net is a novel cross-attention-based loop closure detection framework, used for LiDAR data along with its multi-class information maps and a Cross Transformer module to effectively estimate frame similarity and point cloud overlap~\citep{zheng2024crosstransformer}. A transformer-based image captioning model that enhances visual representations by incorporating the Multi Grid Feature Aggregation mechanism was introduced to integrate multi-level grid features~\citep{bui2024transformer}. Other studies have incorporated spectral and self-attention mechanisms to refine the remote sensing image classification~\citep{qing2021improved}. However, despite their strong performance in spectral feature learning, transformer-based models often struggle with spatial information representation in remote sensing data. To address this, a multi-scale transformer capable of extracting multilevel contextual features was introduced~\citep{tang2022emtcal}. Further advancements in transformer models for remote sensing data fusion include CSMFormer~\citep{gao2023crossscale}, a flexible architecture designed for tasks such as data fusion, land cover classification, and segmentation, which improved the data quality and classification accuracy. Furthermore, a Mixing Self-Attention and Convolution Network~\citep{li2023mixing} is a unified framework that seamlessly integrates self-attention and convolution to achieve both local and global multi-scale perception, enabling comprehensive feature extraction and efficient fusion for LULC classification.

\subsection{Multi-scale Gaussian Smoothing}
Scale-space representations~\citep{witkin1983scale} have been pivotal in feature detection within image processing, advancing multi-scale feature analysis. Multi-scale processing provides greater resilience to noise, effectively extracts coarse features in low-curvature regions, and captures additional structural information, especially feature lines. A multi-scale classification operator can be used for line-type feature extraction in point-sampled geometry, which employs neighborhood sizes as discrete scale parameters~\citep{pauly2003multiscale}, to improve detection accuracy and noise resistance. Scale-space theory pertains to image blurring levels rather than actual image size, with Gaussian filters applied for smoothing~\citep{lowe2004distinctive}. Coarse scales capture shapes and finer scales detect details, such as edges, corners, and textures, that are essential for downstream tasks, \eg~object detection and image matching. The choice of scale parameter depends on the kernel used. For instance, a Gaussian kernel is effective for scale transformations, where for an image function $I(x,y)$, where ($x,y$) is the pixel location, a Gaussian scale space is created by applying Gaussian filters with standard deviations $\sigma$ as scale parameters~\citep{badaud1986uniqueness}. The smoothed images, $L(x, y, \sigma)$, capture details based on scale. The Gaussian pyramid is used for multi-scale image encoding using Gaussian filters, and when combined with a Laplacian pyramid and a CNN model, the Gaussian-Laplacian pyramid performs data compression within a CNN classifier for improving performance for hyperspectral image classification~\citep{chang2024iterative}. Our proposed multi-scale Gaussian smoothing for data fusion is novel for the LCZ classification application.

\section{Our Proposed Models} \label{sec:method}
In this section, we describe our proposed fusion strategies in the CNN-based DL architecture designs for LCZ classification, data-level band grouping strategy, and label merging strategy.

\subsection{Dataset}
The So2Sat LCZ42 dataset~\citep{zhu2020so2sat} consists of Sentinel-1 (S1) SAR images and their corresponding Sentinel-2 (S2) MSI images from 42 regions worldwide. Each image pair is considered as a \textit{scene}, thus, making this a benchmark dataset for scene-level classification (in contrast to pixel- or object-level classification)~\citep{wu2024df4lcz}. The summary description of the dataset is given in the Table ~\ref{tab:so2sat-summary}. S1 and S2 belong to the satellite constellations launched by the European Space Agency's Copernicus program in 2014 and 2015, respectively, with the goal of providing services for emergency management, LULC mapping, climate change monitoring, etc. 

\begin{table}[t]
\centering
\caption{Summary of the So2Sat LCZ42 dataset~\citep{zhu2020so2sat}.}
\label{tab:so2sat-summary}
\begin{tabular}{p{0.28\columnwidth}|p{0.65\columnwidth}}
\hline
\textbf{Specification} & \textbf{Description} \\
\hline
\textbf{Satellite Sensors}
& Sentinel-1 (S1): SAR sensor with 10m spatial, C-band (5.4GHz) spectral resolution, 6-12 days revisit time. \\
& Sentinel-2 (S2): MSI sensor with 10-60m spatial resolution, 13 spectral bands, 5 days revisit time. \\\\
\textbf{Dataset Specifications}
& 400,673 SAR-MSI image (patch) pairs. \\
& 32$\times$32 pixels per image (\ie~320m$\times$320m on the ground). \\
& 8 real-valued bands from VH and VV polarizations of S1 data. \\
& 10 real-valued optical bands of S2 data at 10m spatial resolution. \\
& 42 urban agglomerations and 10 smaller urban areas, worldwide. \\
& \textit{Scene}-level classification \\\\
\textbf{Data Split} 
& Training: 352,366 pairs. \\
for LCZ Classifier Learning 
& Validation: 24,119 pairs. \\
& Testing: 24,188 pairs. \\
\hline
\end{tabular}
\end{table}

\subsection{Fusion Strategies in DL Models}
There are three basic fusion strategies used in DL architectures using multimodal data, namely, \textit{early fusion} (\ie~data-level), \textit{intermediate fusion} (\ie~feature-level), and \textit{late fusion} (\ie~decision-level). \textit{Hybrid} fusion includes multiple basic fusion outputs. Hybrid fusion involving fusion of basic fusion outcomes can also be referred to as \textit{multi-level} fusion. The summary of our proposed CNN-based DL architectures in Figure~\ref{fig:fusionmodels} shows that three models, namely, FM1, FM2, and FM3, are hybrid or multi-level fusion models using early- and intermediate-level fusions, and FM4 uses only late fusion. The four fusion strategies, given as follows, are explained in the subsections:
\begin{enumerate}
  \renewcommand{\labelenumi}{(\roman{enumi})}
  \item FM1 - A \textit{baseline} hybrid fusion model for pixel- and feature-level integration.
  \item FM2 - An attention-based hybrid fusion, which is the FM1 model enhanced with self- and cross-attention mechanisms.
  \item FM3 - A multi-scale hybrid fusion, which is the FM1 model refined with multi-scale Gaussian smoothing of each data modality. 
  \item FM4 - A decision-level fusion model, which uses a weighted combination of CNN and U-Net outputs of SAR and MSI modalities, respectively.
\end{enumerate}

\subsubsection {FM1: Hybrid Fusion}
\begin{figure}[t]
  \centering
  \includegraphics[width=0.9\columnwidth]{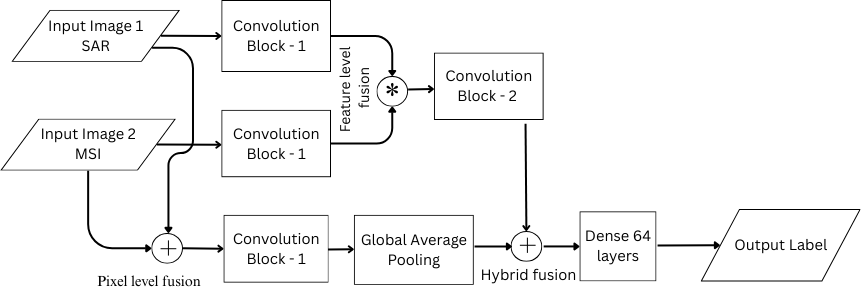} \\
  \vspace{1em}
  \fbox{\includegraphics[width=0.35\columnwidth]{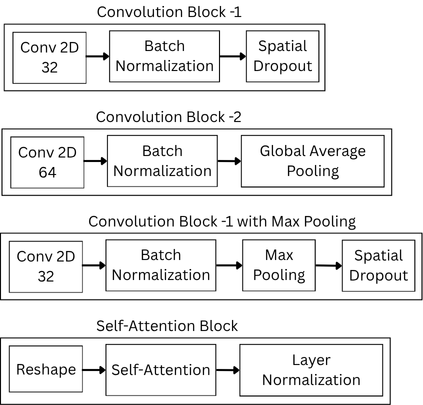}} \\
  \small Convolution Blocks \\
  \vspace{1em}
  \caption{High-level architecture of the FM1 model.}
  \label{fig:fm1}
\end{figure}

Pixel-level fusion directly integrates raw multimodal information, whereas feature-level fusion combines features related to surface characteristics, vegetation structure, building density and height, and anthropogenic thermal emissions to enhance semantic discrimination. Hybrid fusion combines the effects of both.

FM1 separately implements both pixel- and feature-level homogeneous fusion, and then performs a hybrid fusion of the fused maps to utilize complementary information from data and high-level feature representations, as shown in Figure~\ref{fig:fm1}. At the pixel-fusion stage, multisource raw images are concatenated along the spectral channel dimension to generate a unified input array. This fused image is processed using a CNN comprising 32 filters of size 3$\times$3, followed by ReLU activation and batch normalization to enhance discriminative pattern extraction. A spatial dropout with a rate of 0.2 is subsequently applied to reduce overfitting, and the output is aggregated using global average pooling.

For feature extraction, S1 and S2 inputs are independently processed through parallel convolutional blocks, each utilizing 32 filters of size 3$\times$3 with ReLU activation, batch normalization, and spatial dropout. For feature-level fusion, the modality-specific features are fused through element-wise multiplication, yielding an integrated representation. This combined feature map is further refined via additional convolutional layers (64 filters, kernel size 3$\times$3) followed by max-pooling, batch normalization, ReLU activation, and global average pooling.

Finally, the feature maps obtained from both the pixel- and feature-level fusion are concatenated as hybrid (or heterogeneous) fusion, and fed into a fully connected dense layer with 64 neurons. The concatenated feature map is classified by softmax activation into seventeen LCZ classes based on the highest class probability. 

\subsubsection {FM2: Hybrid Fusion with Attention Mechanisms}
\begin{figure}[t]
  \centering    
  \includegraphics[width=0.9\columnwidth]{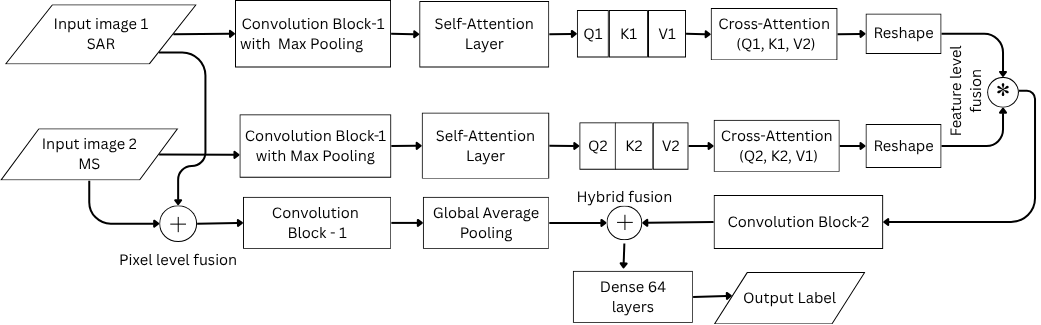} \\
  \vspace{1em}
  \caption{High-level architecture of the FM2 model.}
  \label{fig:fm2}
\end{figure}

As an enhancement to FM1, FM2 incorporates self- and cross-attention mechanisms to model contextual dependencies in the spatial-spectral domain and thereby improves inter-modality feature alignment. The attention mechanisms are implemented in extracted features before the feature-level fusion, as shown in Figure~\ref{fig:fm2}. Feature extraction is done as in FM1, but is followed by max-pooling. The resulting feature maps are reshaped into sequences to be compatible with the attention mechanisms.

For feature enhancement, a self-attention module with eight attention heads is first applied to the feature map of each modality to capture long-range dependencies, refine intra-modal representations, and emphasize contextually informative features. Then, cross-attention modules are added for inter-modal interaction. Specifically, query embeddings derived from S1 features attend to key–value pairs from S2 features, and vice versa, ensuring reciprocal information exchange and improved cross-modal alignment. The outputs of these modules are then reshaped back into their original spatial dimensions to preserve structural coherence.

The pixel-level fusion process in FM2 remains identical to FM1. The attention-enhanced features from both modalities are fused as done in feature-level fusion in FM1. The pixel- and feature-level fused maps are then concatenated as a hybrid fusion step and processed to obtain probability outputs across LCZ classes, as done in FM1. 

\subsubsection{FM3: Hybrid Fusion after Multi-scale Gaussian Smoothing}
\begin{figure}[t]
  \centering
  \includegraphics[width=\columnwidth]{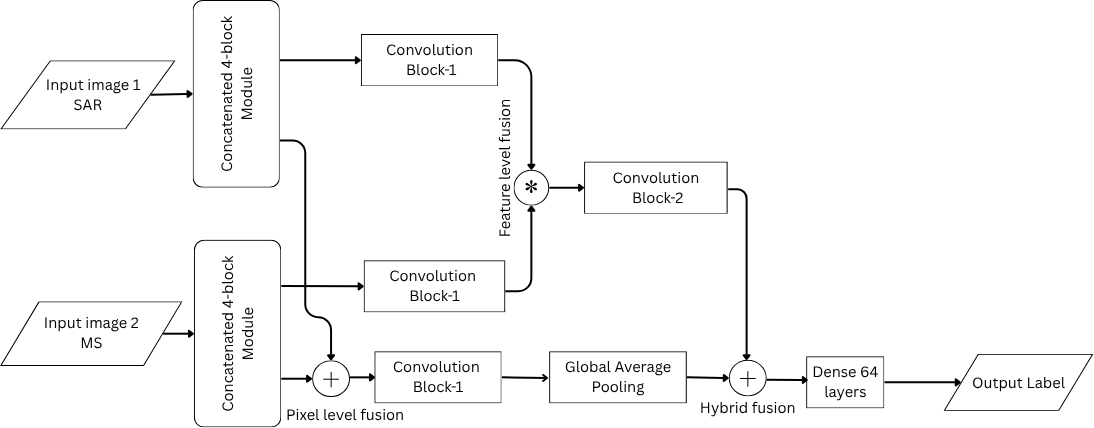} \\
  \vspace{1em}
  \fbox{\includegraphics[width=0.3\columnwidth]{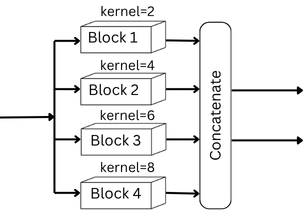}} \\
  \small Concatenated 4-Block Module \\
  \vspace{1em}
  \caption{High-level architecture of the FM3 model.}
  \label{fig:fm3}
\end{figure}

Multi-scale Gaussian smoothing of input images improves the discriminatory nature of spectral and textural characteristics of the raw data. Coarser scales retain shape information, and finer scales retain details, \eg~sharp features and textures. Here, FM3 extends FM1 by processing the image data using multi-scale Gaussian filtering, as shown in Figure~\ref{fig:fm3}. Gaussian filters of kernel sizes of 2, 4, 6, and 8 are independently applied to S1 and S2 inputs, thereby generating multiple smoothed representations of the data. This multi-scale smoothing helps to generate scale-aware representations, reduce noise, capture structural variations at different spatial scales, and enhance feature generalization in the downstream stages. 

The pixel-level fusion in FM3 is the same as that of FM1, except that the raw image data input in FM1 is replaced with the Gaussian-filtered outputs in FM3. Similarly, the feature extraction followed by feature-level fusion is done in the same way as that of FM1, except that the input to these steps is the multi-scale smoothed data instead of the raw data. Finally, a hybrid fusion of fused maps is implemented, as done in FM1.

\subsubsection{FM4: Weighted Decision-level Fusion}
\begin{figure}[t]
  \centering    
  \includegraphics[width=0.8\columnwidth]{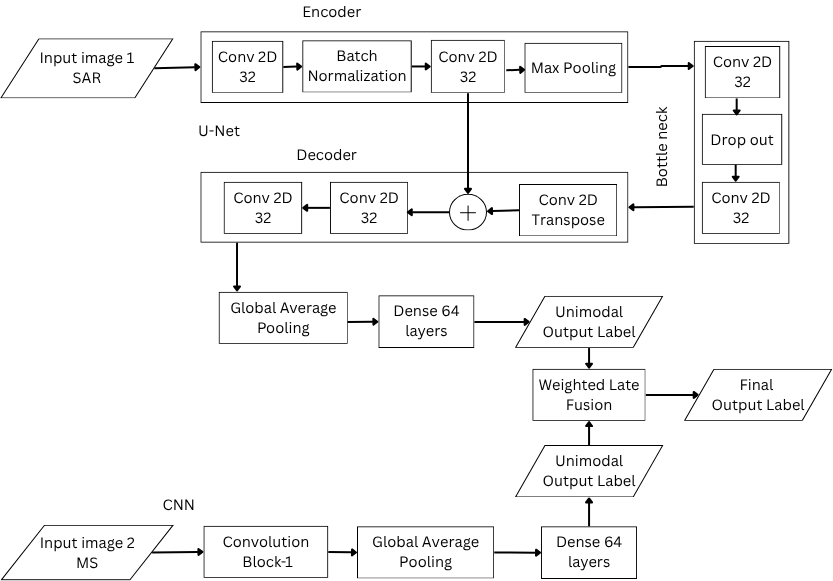} \\
  \vspace{1em}
  \caption{High-level architecture of the FM4 model.}
  \label{fig:fm4}
\end{figure}

Unlike FM1 and its variants, FM4 implements late fusion, \ie~decision-level fusion, as shown in Figure~\ref{fig:fm4}. In FM4, separate DL classifiers are trained for each modality, and their class predictions are combined in a weighted manner at the final stage. For S1 (SAR) data, the U-Net architecture is chosen as the classifier, as it preserves the spatial context and captures hierarchical representations in the image data~\citep{noori2025deep}. For S2 (MSI) data, a vanilla CNN classifier is used, which is optimized for multispectral features~\citep{senecal2019efficient,vasilescu2023cnn}.

Each separate classifier branch yields class probability via a softmax layer. The outputs are then combined through a weighted averaging, where the contribution of each modality is scaled with an empirically determined weight factor, $\alpha$ for U-net and  (1-$\alpha$) for CNN. The weight factor is optimized through tuning experiments to achieve a balance that maximizes classification accuracy. The FM4 model reduces the risk of error propagation from early and intermediate fusion stages, and the optimal weight setting leads to adaptive integration of heterogeneous data sources.

\subsection{Spectral Band Grouping (SBG) Strategy}
SBG has been used in previous studies~\citep{martinez2007clustering,you2022hyperspectral,sawant2021band} to reduce redundancy across spectral bands, and highlight spectrally coherent features. Band selection is guided by the similarity between bands. Spectrally similar bands are clustered using similarity criteria such as correlation measures, distance metrics, or graph-based relationships, and a representative subset is subsequently identified. 

SAR polarization bands exhibit a distinct response to surface properties, while multispectral bands are characterized by their heterogeneous inter-band relationships. To enhance feature extraction from these multi-source datasets, the SAR bands are organized into three groups based on polarization characteristics, whereas the multispectral bands are grouped into four groups according to correlations between bands~\citep{he2023sar}, as shown in Table~\ref{tab:bandgrouping}. The original dataset has eight and ten spectral bands in SAR and MSI images, respectively. The grouped configuration has three groups in SAR images, namely, VH, VV, and CMOE; and four groups in MSI images, namely, RGB, VIR, SWIR, and NIR. The expansions of the group names are given in Table~\ref{tab:bandgrouping}. VIR, SWIR, and NIR are bands related to vegetation, water detection, and less sensitivity to cloud presence, respectively.

These grouped bands are fed into modality-specific CNN branches within the fusion framework, enabling complementary spatial–spectral representation learning. We compare the original and grouped band configurations in our experiments. 

\begin{table}[t]
  \caption{Spectral band grouping (SBG) in SAR-MSI image pairs~\citep{he2023sar} of So2Sat LCZ42 dataset~\citep{zhu2020so2sat}.}
  \label{tab:bandgrouping}
  \centering
  \begin{tabular}{c|l||c|l}
    \hline
    \multicolumn{2}{c||}{Original Bands}
    & \multicolumn{2}{c}{Grouped Bands} \\ \hline
    Band Index & Description
    & Group Name & Description \\\hline

    SAR Band-1 & Real part of unfiltered VH channel
    && SAR Band-1 \\
    SAR Band-2 & Imaginary part of unfiltered VH channel
    & SAR VH & SAR Band-2 \\
    SAR Band-3 & Real part of unfiltered VV channel
    && SAR Band-5 \\
    SAR Band-4 & Imaginary part of unfiltered VV channel
    && \\
    SAR Band-5 & Intensity of refined Lee-filtered VH channel
    && SAR Band-3 \\
    SAR Band-6 & Intensity of refined Lee-filtered VV channel
    & SAR VV & SAR Band-4 \\
    SAR Band-7 & Real part of refined Lee-filtered CMOE
    && SAR Band-6 \\
    SAR Band-8 & Imaginary part of refined Lee-filtered CMOE 
    && \\
    && SAR CMOE & SAR Band-7 \\
    &&& SAR Band-8 \\\hline

    MSI Band B2 & 10m GSD
    && MSI Band B2 \\
    MSI Band B3 & 10m GSD
    &MSI RGB & MSI Band B3 \\
    MSI Band B4 & 10m GSD
    && MSI Band B4\\
    MSI Band B5 & upsampled to 10m from 20m GSD
    && \\
    MSI Band B6 & upsampled to 10m from 20m GSD
    && MSI Band B5 \\
    MSI Band B7 & upsampled to 10m from 20m GSD
    & MSI VRE & MSI Band B6 \\
    MSI Band B8 & 10m GSD
    && MSI Band B7 \\
    MSI Band B8a & upsampled to 10m from 20m GSD
    && MSI Band B8a \\
    MSI Band B11 & upsampled to 10m from 20m GSD
    && \\
    MSI Band B12 & upsampled to 10m from 20m GSD
    & MSI SWIR & MSI Band B11 \\
    &&& MSI Band B12 \\
    &&& \\
    && MSI NIR & MSI Band B8 \\
    \hline
  \end{tabular}
  \begin{center}
    Key --
    CMOE: covariance matrix off-diagonal element ; 
    GSD: ground sampling distance \\
    VRE: vegetation red edge ; 
    SWIR: short-wave infrared ; 
    NIR: near-infrared 
  \end{center}
\end{table}

\subsection{Label Merging (LM) Strategy}
The LCZ standardization distinguishes its seventeen classes based on both urban and natural morphological types~\citep{stewart2012local}. However, many of these classes exhibit strong spectral or structural similarities, often leading to misclassification between adjacent classes in the list (Figure~\ref{fig:dataset}). For instance, \textit{surface albedo} is a spectral property that does not change significantly based on the height of buildings or the density of vegetation~\citep{stewart2012local}. Surface albedo is also captured significantly by MSI in S2 data. It is defined as follows:

\begin{definition}
  \textbf{Surface albedo} is a radiative and spectral property of objects, measured as the ratio of the amount of solar radiation reflected by a surface to the total amount received by it, and is known to vary based on surface properties, such as color, wetness, and roughness.
\end{definition}

Hence, to alleviate the issue of misclassification, we propose LM for LCZ classes, which are similar based on surface albedo and semantics, into eight broad categories, as shown in Table~\ref{tab:lczmergedclass}. The eight merged classes retain the spectral and semantic characterizations in the finer-grained original seventeen LCZ classes. We compare the standard 17-class and the merged 8-class settings in our experiments.

\begin{table}[t]
\caption{LCZ merged class list based on surface albedo and semantics.}
\label{tab:lczmergedclass}
\centering
\begin{tabular}{c|c|c}
\hline
\textbf{Merged Label} & \textbf{Original Labels} & \textbf{Surface Albedo}
\\\hline     
Compact built types
& LCZ 1-3 (compact high/mid/low-rise)
& 0.10-0.20 \\

Open built types
& LCZ 4–6 (open high/mid/low-rise)
& 0.12-0.25 \\

Low-rise built types
& LCZ 7–9 (lightweight/large low-rise, sparsely built)
& 0.15-0.25 \\

Heavy industry
& LCZ 10
& 0.12-0.20 \\

Dense vegetation
& LCZ A-B (dense, scattered trees)
& 0.15-0.20 \\

Low vegetation
& LCZ C–D (bush/grass, low plants)
& 0.15-0.25 \\

Bare surfaces
& LCZ E-F (bare rock/paved, bare soil/land)
& 0.20-0.30 \\

Water
& LCZ G
& 0.02-0.10 \\
\hline

\end{tabular}
\end{table}

\subsection{Performance Metrics}
Model performance is assessed using Accuracy,  Precision, Recall, F1 Score~\citep{sokolova2009systematic}, Overall Accuracy, and Kappa Coefficient, $\kappa$~\citep{cohen1960coefficient}, at the average class level and per class. These metrics give a balanced evaluation across both majority and minority classes, and a direct comparison between fusion strategies.

The widely used class-wise performance metrics in machine learning are computed with $TP$, $TN$, $FP$, and $FN$, which are the number of true positives, true negatives, false positives, and false negatives, respectively, for each class. \textit{Accuracy} determines the proportion of correctly classified instances over the total number of instances. \textit{Precision} is the proportion of true positives over the total number of predicted positives. Recall is the proportion of true positives over the total number of actual positives. \textit{F1 Score} is the harmonic mean of precision and recall. Multi-class classifiers are assessed using \textit{Overall Accuracy}, which is the average of accurate predictions of $k$ classes, with $TP_i$ true positives in $i^{th}$ class and a total of $N$ samples.

Kappa coefficient ($\kappa$) is another widely used metric that measures the inter-rater reliability between the true and predicted labels, using $P_o$ and $P_e$ that represent the proportion of observed agreement between raters, and the proportion of agreement expected by chance, respectively.
\begin{eqnarray}
\text{Accuracy }& A &= \frac{TP + TN}{TP + TN + FP + FN} \nonumber \\
\text{Precision }& P &= \frac{TP}{TP + FP} \nonumber \\
\text{Recall }& R &= \frac{TP}{TP + FN} \nonumber \\
\text{F1 Score }& F_1 &= 2\times\frac{P\cdot R}{P+R} \nonumber \\
\text{Overall Accuracy }& OA &= \frac{1}{N}\sum\limits_{i=1}^k TP_i \nonumber \\
\text{Kappa Coefficient }& \kappa &= \frac{P_o - P_e}{1 - P_e} \nonumber
\end{eqnarray}

The macro- and weighted averages of the metric $M$ using class-wise metric values, \eg~$P$, $R$, $F_1$, and $\kappa$, for $K$ classes, with $N_k$ and $N$ as class-wise support for $k$\textsuperscript{th} and total support, respectively, are computed as follows:
\begin{eqnarray}
\text{Macro-average }& \mu(M) &= \frac{\sum\limits_{k=1}^K M_k}{K} \nonumber \\
\text{Weighted average }& \mu_w(M) &=\sum\limits_{k=1}^K \frac{N_k}{N}\cdot M_i \nonumber
\end{eqnarray}

In addition to these metrics, we also use a confusion matrix for assessing class-wise performance in LCZ classification. For LM, we ensure to handle \textbf{internal class-to-class errors} in the confusion matrix when converting a matrix for original labels to one for merged labels. This is used in our validation with the state-of-the-art (SOTA) models with the original 17 classes, which are recomputed for the 8 merged classes (given in Table~\ref{tab:lczmergedclass}) for comparing with our models with LM.

Given the dataset has imbalanced classes, we compute the Matthews Correlation Coefficient (MCC), which is the most robust measure for binary classification~\citep{chicco2020advantages}, extensible to multiclass classification. 
\begin{eqnarray}
\text{Matthews Correlation Coefficient }& MCC &= \frac{c.s-\sum\limits_{k=1}^K p_k.t_k}{\sqrt{\big(s^2-\sum\limits_{k=1}^K p_k^2\big).\big(s^2-\sum\limits_{k=1}^K t_k^2\big)}}, \nonumber
\end{eqnarray}
\begin{eqnarray}
\text{where }& c &=\sum\limits_{k=1}^K TP_k \text{, total number of correctly predicted samples}\nonumber \\
& s &= N \text{, total support} \nonumber \\
& t_k &= TP_k+FN_k \text{, total number of actual instances of class }k \nonumber \\
& p_k &= TP_k+FP_k \text{, total number of predicted instances of class }k \nonumber
\end{eqnarray}

\section{Experimental Design} \label{sec:experiments}
All models are tested, and experiments are conducted on the So2Sat LCZ42 dataset. Though each model exhibits unique architectural characteristics, we use the same setup for their implementation to enable a fair comparison of outputs. We also perform a systematic ablation study of the proposed models and strategies.

\subsection{Implementation}   
All models are implemented in Python using TensorFlow and the Keras framework. The input dataset is split into training, validation, and test data. The fusion models are trained using identical data splits across fusion strategies and an ablation study. Table~\ref{tab:hyperparam} shows the optimal values of hyperparameters obtained for the CNN fusion model. Model training is performed on the same hardware configuration for all models and their ablation experiments, on a high-memory compute node of 2$\times$ Intel Xeon Cascade Lake 8268, 24 cores, 2.9 GHz, processors with 768 GB memory. 

\begin{table}[t]
  \centering
  \caption{Optimized hyperparameters for our implementation of models FM1-FM4 and their ablation experiments.}
  \label{tab:hyperparam}
  \begin{tabular}{l|ccccc}
    \hline
    Hyperparameters & Optimizer & Learning & \# Epochs & Spatial & Loss \\
    && Rate && Dropout Rate & Function \\\hline
    Values & Adam & 0.0001 & 100 & 0.2 & Categorical \\
    &&&&& Cross Entropy \\\hline
  \end{tabular}

  \vspace{1em}

  \caption{Model settings for systematic experimentation on the proposed FM1-FM4 models, including an ablation study.}
  \label{tab:notation}
  \begin{adjustbox}{width=0.95\columnwidth}
  \begin{tabularx}{\columnwidth}{l|cccc|cc|l}
    \hline
    Model & P.-level & F.-level & Hybrid & D.-level
    & Attn. & MSG
    & Remarks \\
    & Fusion & Fusion & Fusion & Fusion
    & Mech. & Smth.
    & \\\hline

    \multicolumn{8}{c}{\bf FM* Experiments (including Ablation Study)} \\\hline
    FM1 & $\checkmark$ & $\checkmark$ & $\checkmark$ &
    &&& Baseline hybrid model \\

    FM1a & $\checkmark$ &&&
    &&& Early fusion model \\

    FM1b && $\checkmark$ &&
    &&& Intermediate fusion model \\
    \hline

    FM2 & $\checkmark$ & $\checkmark$ & $\checkmark$ &
    & $\checkmark$ &
    & FM1 with attention \\

    FM2b && $\checkmark$ &&
    & $\checkmark$ &
    & FM1b with attention \\
    \hline

    FM3 & $\checkmark$ & $\checkmark$ & $\checkmark$ &
    && $\checkmark$
    & FM1 with multi-scale data \\

    FM3a & $\checkmark$ &&& 
    && $\checkmark$
    & FM1a with multi-scale data \\

    FM3b && $\checkmark$ &&
    && $\checkmark$
    & FM1b with multi-scale data \\
    \hline

    FM4 &&&& $\checkmark$
    &&
    & Weighted late fusion \\
    \hline
    
    \multicolumn{8}{c}{\bf Experiments with Grouping Strategies (including Ablation Study)} \\\hline
    FM*\textsubscript{B} &&&&
    &&
    & FM* with SBG \\
    FM*\textsubscript{L} &&&&
    &&
    & FM* with LM \\
    FM*\textsubscript{BL} &&&&
    &&
    & FM* with SBG and LM \\
    \hline
  \end{tabularx}
  \end{adjustbox}
  \vspace{0.25em}
  \begin{center}
    P.-level: Pixel-level; F.-level: Feature-level; D.-level: Decision-level;\\
    Attn. Mech.: Attention Mechanism;
    MSG Smth.: Multi-scale Gaussian Smoothing;\\
    SBG: Spectral Band Grouping; LM: Label Merging. \\
  \end{center}
\end{table}

A comprehensive ablation study is conducted to determine the contributions of the fusion strategy. Each fusion model is evaluated under pixel-level, feature-level, and hybrid fusion, with and without band grouping.

\subsection{Ablation Experiments}
The influence of early and intermediate fusion, and that of spectral band grouping and label merging are studied using ablation experiments. A few of these ablation experiments are simple fusion models, \eg~early and intermediate fusion models. The notations used for all experiments, including the ablation study and comparison of the SOTA models, \ie~MsF-LCZ-Net~\citep{he2023sar} and MSCA-Net~\citep{liu2025enhancing}, are explained in Table~\ref{tab:notation}.

\section{Results and Validation} \label{sec:results}
We discuss the overall performance of the proposed models, including the fusion and grouping strategies, along with their ablation experiments and class-wise performances. FM1 and FM3 are better-performing hybrid fusion models.

\subsection{Overall Performance of Hybrid Fusion Models}\label{sec:results-overall}
\begin{figure}[tbp]
    \centering
    \includegraphics[width=.8\columnwidth]{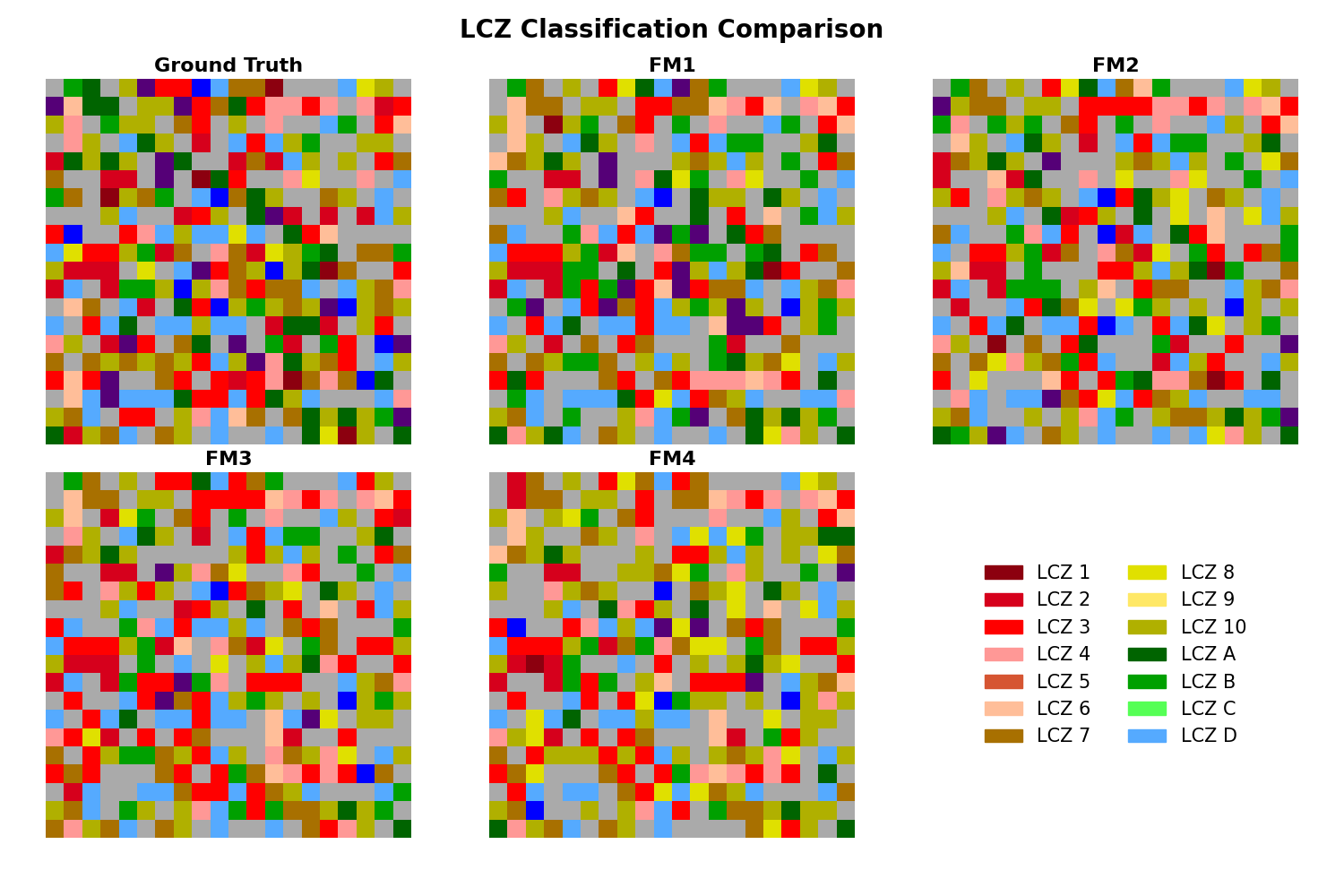}
    \caption{Spatial grid visualization of LCZ labels for the original ground truth and classification outputs arranged in a 20$\times$20 grid (first 400 patches), of FM1--FM4 models.}
    \label{fig:spatial-fusion}
    \vspace{1em}
    \centering
    \captionof{table}{Overall performance metrics of FM1--FM4 models with fusion and grouping strategies ({\color{blue}blue}, {\color{red}red}, {\color{orange}orange} for top 3 results, in descending rank).}
  \label{tab:overall}
  \begin{tabularx}{\columnwidth}{c|ccccc}
    \hline
    Model\textbackslash Metric &
    Overall Accuracy & Precision & Recall & F1 Score & Kappa Coeff. \\
    & $OA$ & $P$ & $R$ & $F_1$ & $\kappa$ \\
    \hline
      FM1 & 0.662 & 0.602 & 0.556 & 0.564 & 0.630 \\
      FM2 & 0.660 & 0.567 & 0.554 & 0.546 & 0.628 \\
      FM3 & 0.655 & 0.592 & 0.540 & 0.544 & 0.620 \\
      FM4 & 0.581 & 0.490 & 0.457 & 0.454 & 0.540 \\
      \hline

      FM1\textsubscript{B} & 0.681 & 0.612 & 0.560 & 0.558 & 0.650 \\
      FM2\textsubscript{B} & 0.637 & 0.576 & 0.549 & 0.543 & 0.603 \\
      FM3\textsubscript{B} & 0.665 & 0.587 & 0.553 & 0.550 & 0.633 \\
      FM4\textsubscript{B} & 0.624 & 0.525 & 0.475 & 0.473 & 0.587 \\
      \hline
      
      FM1\textsubscript{L} & \color{orange}0.737 & \color{orange}0.708 & \color{red}0.735 & \color{orange}0.713 & \color{orange}0.689 \\
      FM2\textsubscript{L} & 0.731 & 0.698 & 0.725 & 0.707 & 0.682 \\
      FM3\textsubscript{L} & 0.731 & 0.697 & 0.718 & 0.704 & 0.682 \\
      FM4\textsubscript{L} & 0.699 & 0.643 & 0.638 & 0.637 & 0.600 \\
      \hline

      FM1\textsubscript{BL} & \color{blue}0.766 & \color{blue}0.740 & \color{blue}0.754 & \color{blue}0.744 & \color{blue}0.723 \\
      FM2\textsubscript{BL} & 0.720 & 0.693 & 0.716 & 0.700 & 0.669 \\
      FM3\textsubscript{BL} & \color{red}0.751 & \color{red}0.724 & \color{orange}0.729 & \color{red}0.723  & \color{red}0.705 \\
      FM4\textsubscript{BL} & 0.699 & 0.643 & 0.663 & 0.650 & 0.644 \\
      \hline
  \end{tabularx}
\end{figure}

Figure~\ref{fig:spatial-fusion} shows the visualization of labels of the first 400 patches in the dataset laid out as a 20$\times$20 grid, owing to the absence of availability of geolocation in the dataset. We observe that none of the proposed fusion models, FM1, FM2, FM3, and FM4, perform perfectly to generate results similar to those of the ground truth (GT).

While these 400 patches need not be the best representation of the model performance, we use different metrics, given in Table~\ref{tab:overall}, to compare the overall performance of these models, and their variants from grouping strategies. Our observations include:
\begin{itemize}
\item FM1 outperforms the other three models, and FM3 is the second best. In the descending order of effectiveness, (i) band grouping and label merging, (ii) label merging alone, and (iii) band grouping alone, outperform the vanilla models.
\item FM1 uses two levels of homogeneous fusion, and one level of hybrid fusion, which enable in better spatial-spectral information extraction, and FM3 uses multiple scales for data transformation, which further follows the FM1 model. Here, FM3 is not as effective as FM1, which could be because the current FM3 implementation uses static scale settings.
\item Label merging absorbs intra-class misclassifications within a merged label, thus improving accuracy metrics.
\item Without considering label merging, we observe that band grouping improves the performance of the hybrid fusion models. FM3 could be further improved using adaptive or dynamic scale settings.
\item Late fusion, \ie~FM4 and its variants, consistently underperforms, which reinforces that early and intermediate fusion improves the performance.
\item We also observe that the widely used performance metrics (\ie~$OA$, $P$, $R$, $F_1$, $\kappa$) behave consistently across models.
\end{itemize}
Thus, the top three hybrid models are FM1\textsubscript{BL}, FM3\textsubscript{BL}, and FM1\textsubscript{L}.

The ablation experiments show that models with simpler fusion mechanisms, \eg~early fusion in FM1a and FM3a, are more efficient compared to other hybrid fusion models, as observed in Table~\ref{tab:time}. Attention mechanisms, \ie~in FM2 and FM2b, make the models extremely inefficient. The timing measurements are similar for the variants with grouping mechanisms (\ie~ FM*\textsubscript{B}, FM*\textsubscript{L}, FM*\textsubscript{BL}), given in Table~\ref{tab:time}. Here, the FM1 model and its variants are considered to be a good trade-off between computational complexity and accuracy.

In Table~\ref{tab:ablation}, we record the performance of models in ablation experiments using metrics commonly used in the comparisons with the SOTA, namely, the overall accuracy metrics for all classes $OA$, for built-up classes (\ie~Classes 1 to 10) $OA_{bu}$, and for natural classes (\ie~Classes A to G) $OA_n$, and Kappa coefficient $\kappa$. We again observe that FM1\textsubscript{BL} outperforms other models, and its variant with early fusion of data alone, \ie~FM1b\textsubscript{BL}, performs the next best. Table~\ref{tab:sota-comparison} reinforces the best performance of FM1\textsubscript{BL}, as it outperforms the SOTA models.

Additional results are given in Appendix~\ref{app:additional-results} in Table~\ref{tab:additional-results}, for performance metrics of $P$ and $R$ for all ablation experiments given in Table~\ref{tab:ablation}.

\begin{table}[tp]
  \centering
  \caption{Training time of ablation experiments in hours ({\color{blue}blue}, {\color{red}red}, {\color{orange}orange} for top 3 results, in descending rank).}
  \label{tab:time}
  \begin{tabular}{ccc|cc|ccc|c}
    \hline
    FM1 & FM1a & FM1b
    & FM2 & FM2b
    & FM3 & FM3a & FM3b
    & FM4 \\
    \hline
    3.489 & \color{blue} 1.289 & \color{orange} 3.031
    & 27.406 & 27.081
    & 4.141 & \color{red} 2.033 & 3.513
    & 4.302 \\
    \hline
  \end{tabular}
  \vspace{1em}
  \centering
  \caption{Performance metrics of ablation experiments with fusion and grouping strategies ({\color{blue}blue}, {\color{red}red}, {\color{orange}orange} for top 3 results, in descending rank) for each metric.}
  \label{tab:ablation}
  \begin{adjustbox}{width=0.8\columnwidth}
  \begin{tabularx}{\columnwidth}{ccc|cc|ccc|c}
    \hline
    \multicolumn{9}{c}{\bf Overall Accuracy $OA$} \\
    \hline
    FM1 & FM1a & FM1b
    & FM2 & FM2b
    & FM3 & FM3a & FM3b
    & FM4 \\
    0.662 & 0.608 & 0.642
    & 0.660 & 0.623
    & 0.655 & 0.579 & 0.637
    & 0.581 \\
    \hline
    FM1\textsubscript{B} & FM1a\textsubscript{B} & FM1b\textsubscript{B}
    & FM2\textsubscript{B} & FM2b\textsubscript{B}
    & FM3\textsubscript{B} & FM3a\textsubscript{B} & FM3b\textsubscript{B}
    & FM4\textsubscript{B} \\
    0.681 & 0.591 & 0.672 
    & 0.637 & 0.647
    & 0.665 & 0.666 & 0.656
    & 0.624 \\
    \hline
    FM1\textsubscript{L} & FM1a\textsubscript{L} & FM1b\textsubscript{L}
    & FM2\textsubscript{L} & FM2b\textsubscript{L}
    & FM3\textsubscript{L} & FM3a\textsubscript{L} & FM3b\textsubscript{L}
    & FM4\textsubscript{L} \\
    0.737 & 0.699 & 0.726
    & 0.731 & 0.702
    & 0.731 & 0.653 & 0.722
    & 0.664 \\
    \hline
    FM1\textsubscript{BL} & FM1a\textsubscript{BL} & FM1b\textsubscript{BL}
    & FM2\textsubscript{BL} & FM2b\textsubscript{BL}
    & FM3\textsubscript{BL} & FM3a\textsubscript{BL} & FM3b\textsubscript{BL}
    & FM4\textsubscript{BL} \\
    \color{blue}0.766 & 0.670 & \color{red}0.759
    & 0.720 & 0.734
    & \color{orange}0.751 & 0.743 & 0.742
    & 0.699 \\
    \hline

    \hline
    \multicolumn{9}{c}{\bf Overall Accuracy (Built-up Classes) $OA_{bu}$} \\
    \hline
    FM1 & FM1a & FM1b
    & FM2 & FM2b
    & FM3 & FM3a & FM3b
    & FM4 \\
    0.590 & 0.524 & 0.561
    & 0.571 & 0.529
    & 0.573 & 0.500 & 0.565
    & 0.510 \\
    \hline
    FM1\textsubscript{B} & FM1a\textsubscript{B} & FM1b\textsubscript{B}
    & FM2\textsubscript{B} & FM2b\textsubscript{B}
    & FM3\textsubscript{B} & FM3a\textsubscript{B} & FM3b\textsubscript{B}
    & FM4\textsubscript{B} \\
    0.620 & 0.520 & 0.596
    & 0.565 & 0.566
    & 0.603 & 0.613 & 0.588
    & 0.543 \\
    \hline
    FM1\textsubscript{L} & FM1a\textsubscript{L} & FM1b\textsubscript{L}
    & FM2\textsubscript{L} & FM2b\textsubscript{L}
    & FM3\textsubscript{L} & FM3a\textsubscript{L} & FM3b\textsubscript{L}
    & FM4\textsubscript{L} \\
    0.644 & 0.599 & 0.632
    & 0.631 & 0.587
    & 0.638 & 0.562 & 0.632
    & 0.570 \\
    \hline
    FM1\textsubscript{BL} & FM1a\textsubscript{BL} & FM1b\textsubscript{BL}
    & FM2\textsubscript{BL} & FM2b\textsubscript{BL}
    & FM3\textsubscript{BL} & FM3a\textsubscript{BL} & FM3b\textsubscript{BL}
    & FM4\textsubscript{BL} \\
    \color{blue}0.685 & 0.604 & \color{orange}0.676
    & 0.633 & 0.635
    & \color{red}0.683 & 0.672 & 0.657
    & 0.608 \\
    \hline

    \hline
    \multicolumn{9}{c}{\bf Overall Accuracy (Natural Classes) $OA_n$} \\
    \hline
    FM1 & FM1a & FM1b
    & FM2 & FM2b
    & FM3 & FM3a & FM3b
    & FM4 \\
    0.770 & 0.732 & 0.763
    & 0.792 & 0.763
    & 0.776 & 0.696 & 0.745
    & 0.688 \\
    \hline
    FM1\textsubscript{B} & FM1a\textsubscript{B} & FM1b\textsubscript{B}
    & FM2\textsubscript{B} & FM2b\textsubscript{B}
    & FM3\textsubscript{B} & FM3a\textsubscript{B} & FM3b\textsubscript{B}
    & FM4\textsubscript{B} \\
    0.772 & 0.697 & 0.785
    & 0.743 & 0.768
    & 0.756 & 0.744 & 0.758
    & 0.746 \\
    \hline
    FM1\textsubscript{L} & FM1a\textsubscript{L} & FM1b\textsubscript{L}
    & FM2\textsubscript{L} & FM2b\textsubscript{L}
    & FM3\textsubscript{L} & FM3a\textsubscript{L} & FM3b\textsubscript{L}
    & FM4\textsubscript{L} \\
    0.875 & 0.847 & 0.867
    & 0.879 & 0.873
    & 0.871 & 0.787 & 0.857
    & 0.804 \\
    \hline
    FM1\textsubscript{BL} & FM1a\textsubscript{BL} & FM1b\textsubscript{BL}
    & FM2\textsubscript{BL} & FM2b\textsubscript{BL}
    & FM3\textsubscript{BL} & FM3a\textsubscript{BL} & FM3b\textsubscript{BL}
    & FM4\textsubscript{BL} \\
    \color{blue}0.888 & 0.770 & \color{red}0.884
    & 0.848 & \color{orange}0.881
    & 0.851 & 0.847 & 0.868
    & 0.835 \\
    \hline
    
    \hline
    \multicolumn{9}{c}{\bf Kappa Coefficient $\kappa$} \\
    \hline
    FM1 & FM1a & FM1b
    & FM2 & FM2b
    & FM3 & FM3a & FM3b
    & FM4 \\
    0.630 & 0.580 & 0.609
    & 0.628 & 0.588
    & 0.620 & 0.538 & 0.603
    & 0.540 \\
    \hline
    FM1\textsubscript{B} & FM1a\textsubscript{B} & FM1b\textsubscript{B}
    & FM2\textsubscript{B} & FM2b\textsubscript{B}
    & FM3\textsubscript{B} & FM3a\textsubscript{B} & FM3b\textsubscript{B}
    & FM4\textsubscript{B} \\
    0.650 & 0.550 & 0.640
    & 0.603 & 0.614
    & 0.633 & 0.633 & 0.622
    & 0.587 \\
    \hline
    FM1\textsubscript{L} & FM1a\textsubscript{L} & FM1b\textsubscript{L}
    & FM2\textsubscript{L} & FM2b\textsubscript{L}
    & FM3\textsubscript{L} & FM3a\textsubscript{L} & FM3b\textsubscript{L}
    & FM4\textsubscript{L} \\
    0.689 & 0.643 & 0.675
    & 0.682 & 0.648
    & 0.682 & 0.588 & 0.670
    & 0.600 \\
    \hline
    FM1\textsubscript{BL} & FM1a\textsubscript{BL} & FM1b\textsubscript{BL}
    & FM2\textsubscript{BL} & FM2b\textsubscript{BL}
    & FM3\textsubscript{BL} & FM3a\textsubscript{BL} & FM3b\textsubscript{BL}
    & FM4\textsubscript{BL} \\
    \color{blue}0.723 & 0.611 & \color{red}0.715
    & 0.669 & 0.685
    & \color{orange}0.705 & 0.695 & 0.694
    & 0.644 \\
    \hline    
  \end{tabularx}
  \end{adjustbox}


  \vspace{1em}
        
  \caption{Comparison of our best result with SOTA LCZ classifiers ({\color{blue}blue}, {\color{red}red}, {\color{orange}orange} for top 3 results, in descending rank).}
  \label{tab:sota-comparison}
  \begin{tabular}{l|cccc}
    \hline
    Model & $OA$ & $OA_{bu}$ & $OA_n$  & $\kappa$ \\
    \hline
    ResNet50~\citep{xie2017aggregated}
    & 0.614 & 0.531 & 0.737 & 0.576 \\
    ViT~\citep{dosovitskiy2020image}
    & 0.595 & 0.534 &0.701 & 0.555 \\
    MSMLA-Net~\citep{kim2021local}
    & 0.662 & 0.638 & 0.735 & 0.630 \\
    Sen2LCZ-Net-MF~\citep{qiu2020multilevel}
    & 0.694 & 0.647 & 0.763 & 0.664 \\
    MsF-LCZ-Net~\citep{he2023sar} 
    & 0.679 & 0.634* & 0.712* & 0.648 \\
    MSCA-Net~\citep{liu2025enhancing}
    & \color{orange} 0.710 & \color{orange}0.657 & \color{orange}0.785 & \color{orange}0.691 \\
    MSCA-MSLCZNet~\citep{liu2025enhancing}
    & \color{red}0.736  & \color{red}0.677 & \color{red}0.823 & \color{red}0.710 \\
    \hline
    \bf Ours (FM1\textsubscript{BL})
    & \bf \color{blue}0.766 & \bf \color{blue}0.685 & \bf \color{blue}0.888 & \bf \color{blue}0.723 \\
    \hline
  \end{tabular}

*The processed metrics are computed from the published class-wise $P$ and $R$.
\end{table}

\subsection{Class-wise Performance of Hybrid Fusion Models}\label{sec:results-classwise}
When we do an in-depth class-wise performance analysis of the models in Table~\ref{tab:classwise-top}, we use the SOTA models, MsF-LCZ-Net (\textbf{M1})~\citep{he2023sar} and MSCA-Net (\textbf{M2})~\citep{liu2025enhancing}. These models have been chosen, as their works have published class-wise performances. It must be noted that \textbf{M2} focuses on using MSI data alone, and is not a fusion model. 

MSCA-Net consistently performs marginally better than our best performing models after band grouping and label merging, \ie~FM1\textsubscript{BL} and FM3\textsubscript{BL}, for all classes, except for class 10 (``Heavy Industry'') and the merged class of classes E and F (``Bare Surfaces''). In Table~\ref{tab:classwise-no-lm}, where we consider the best performing 17-class hybrid fusion classifiers, namely, FM1, FM1\textsubscript{B}, FM3, and FM3\textsubscript{B} (from Table~\ref{tab:overall}), we observe that FM1 outperforms the SOTA models for class 7 (``Lightweight Low-Rise''), class E (``Bare Rock''), and class F (``Bare Soil''); and FM3\textsubscript{B} outperforms for class 1 (``Compact High-Rise'') and class 2 (``Compact Mid-Rise''). We also observe that FM1\textsubscript{B} performs consistently well across all classes, comparable to the SOTA models. These inferences indicate that there are differences in class-wise performance with and without label merging, except for the ``Bare Surfaces'' class. These patterns have also been observed using the metrics $F_1$, $\kappa$, $P$, and $R$.

However, we see that the dataset has a high level of class imbalance (Figure~\ref{fig:dataset}). Hence, it is imperative to check the performance using MCC as a metric. Our results in Table~\ref{tab:classwise-mcc} show that FM1\textsubscript{B} and FM1\textsubscript{BL} perform comparably to the SOTA models, for 17- and 8-class scenarios, \ie~without and with label merging, respectively. Our models FM*\textsubscript{B} and FM*\textsubscript{BL} use band grouping, inspired by \textbf{M1}. Hence, the better performance of our models than MsF-LCZ-Net indicates that hybrid fusion improves the discrimination of features, leading to better classification.

\begin{table}[tp]
  \centering
  \caption{Performance results of our top two hybrid fusion models (from Table~\ref{tab:overall}), and SOTA models, \textbf{MsF-LCZ-Net}~\citep{he2023sar} (\textbf{M1}) and \textbf{MSCA-Net}~\citep{liu2025enhancing} (\textbf{M2}), using selected metrics, along with their averages, \ie~macro ($\mu$) and weighted ($\mu_w$) ({\color{blue}blue}, {\color{red}red}, and {\color{orange}orange} for top 3 results, in descending rank). Since all the top models use LM, the labels show the original classes used for merging.}
  \label{tab:classwise-top}
  \begin{tabular}{c|cccc|cccc}
    \hline
    Label
    & FM1\textsubscript{BL} & FM3\textsubscript{BL}
    & \textbf{M1}\textsuperscript{*}
    & \textbf{M2}\textsuperscript{*} 
    & FM1\textsubscript{BL} & FM3\textsubscript{BL}
    & \textbf{M1}\textsuperscript{*}
    & \textbf{M2}\textsuperscript{*} \\
    \hline
    \rule{0pt}{3ex}    
    & \multicolumn{4}{c|}{F1 Score $F_1$}
    & \multicolumn{4}{c}{Kappa Coefficient $\kappa$} \\
    \hline
    \{1,2,3\}
    & \color{orange}0.742 & \color{red}0.744 & 0.676 & \color{blue}0.768 
    & \color{red}0.693 & \color{orange}0.692 & 0.617 & \color{blue}0.723 
    \\
    \{4,5,6\}
    & \color{red}0.667 & \color{orange}0.661 & 0.653 & \color{blue}0.700 
    & \color{red}0.605 & \color{orange}0.598 & 0.597 & \color{blue}0.642 
    \\
    \{7,8,9\}
    & \color{orange}0.730 & 0.718 & \color{red}0.735 & \color{blue}0.743 
    & \color{red}0.645 & 0.637 & \color{orange}0.643 & \color{blue}0.669 
    \\
    \{10\}
    & \color{red}0.481 & 0.463 & \color{blue}0.501 & \color{orange}0.469 
    & \color{red}0.462 & 0.445 & \color{blue}0.483 & \color{orange}0.446 
    \\
    \{A,B\}
    & \color{orange}0.874 & 0.853 & \color{red}0.889 & \color{blue}0.919 
    & \color{orange}0.857 & 0.834 & \color{red}0.874 & \color{blue}0.908 
    \\
    \{C,D\}
    & \color{orange}0.803 & 0.762 & \color{red}0.844 & \color{blue}0.870 
    & \color{orange}0.765 & 0.716 & \color{red}0.816 & \color{blue}0.847 
    \\
    \{E,F\}
    & \color{blue}0.672 & \color{red}0.598 & \color{orange}0.589 & 0.566 
    & \color{blue}0.660 & \color{red}0.583 & \color{orange}0.574 & 0.552 
    \\
    \{G\}
    & \color{orange}0.987 & 0.981 & \color{red}0.989 & \color{blue}0.990 
    & \color{orange}0.986 & 0.979 & \color{red}0.987 & \color{blue}0.989 
    \\
    $\mu$
    & \color{red}0.744 & 0.723 & \color{orange}0.735 & \color{blue}0.753 
    & \color{red}0.710 & 0.686 & \color{orange}0.699 & \color{blue}0.722 
    \\
    $\mu_w$
    & \color{red}0.766 & 0.750 & \color{orange}0.761 & \color{blue}0.790 
    & \color{red}0.720 & 0.702 & \color{orange}0.713 & \color{blue}0.750 
    \\
    \hline
    \hline
    \rule{0pt}{3ex}    
    & \multicolumn{4}{c|}{Precision $P$}
    & \multicolumn{4}{c}{Recall $R$} \\
    \hline
    \{1,2,3\}
    & \color{red}0.765 & \color{orange}0.721 & 0.717 & \color{blue}0.768 
    & \color{red}0.720 & \color{blue}0.769 & 0.640 & \color{blue}0.769 
    \\
    \{4,5,6\}
    & \color{orange}0.621 & 0.618 & \color{blue}0.692 & \color{red}0.631 
    & \color{red}0.721 & \color{orange}0.710 & 0.618 & \color{blue}0.786 
    \\
    \{7,8,9\}
    & \color{orange}0.794 & \color{red}0.800 & 0.707 & \color{blue}0.821 
    & \color{orange}0.675 & 0.652 & \color{blue}0.765 & \color{red}0.679 
    \\
    \{10\}
    & \color{orange}0.510 & \color{red}0.530 & \color{blue}0.538 & 0.414 
    & \color{orange}0.455 & 0.412 & \color{red}0.468 & \color{blue}0.542 
    \\
    \{A,B\}
    & 0.875 & \color{red}0.900 & \color{orange}0.887 & \color{blue}0.950 
    & \color{orange}0.873 & 0.810 & \color{blue}0.892 & \color{red}0.889 
    \\
    \{C,D\}
    & \color{orange}0.756 & 0.709 & \color{red}0.822 & \color{blue}0.865 
    & \color{orange}0.855 & 0.822 & \color{red}0.868 & \color{blue}0.874 
    \\
    \{E,F\}
    & \color{blue}0.618 & \color{orange}0.546 & 0.534 & \color{red}0.565 
    & \color{blue}0.736 & \color{red}0.660 & \color{orange}0.658 & 0.568 
    \\
    \{G\}
    & \color{orange}0.978 & 0.967 & \color{red}0.981 & \color{blue}0.984 
    & \color{red}0.996 & 0.995 & \color{blue}0.997 & \color{red}0.996 
    \\
    $\mu$
    & \color{red}0.740 & 0.724 & \color{orange}0.735 & \color{blue}0.750 
    & \color{red}0.754 & 0.729 & \color{orange}0.738 & \color{blue}0.763 
    \\
    $\mu_w$
    & \color{red}0.770 & 0.758 & \color{orange}0.762 & \color{blue}0.800 
    & \color{red}0.766 & 0.751 & \color{orange}0.763 & \color{blue}0.787 
    \\
    \hline
  \end{tabular}

*The label-merged results are computed from the published confusion matrix.
\end{table}

\begin{table}[tp]
  \centering
  \caption{F1 score for best-performing models without LM (from Table~\ref{tab:overall}) and SOTA models (used in Table~\ref{tab:classwise-top}), with macro- ($\mu$) and weighted ($\mu_w$) averages. ({\color{blue}blue}, {\color{red}red}, and {\color{orange}orange} for the top 3 results, in descending rank in each class).}
  \label{tab:classwise-no-lm}
  \begin{tabular}{c|cccc|cc}
    \hline
    Class
    & FM1 & FM1\textsubscript{B}
    & FM3 & FM3\textsubscript{B}
    & \textbf{M1}\textsuperscript{*}
    & \textbf{M2}\textsuperscript{*} \\
    \hline
    1
    & 0.411 & 0.422
    & \color{red}0.431 & \color{blue}0.473
    & \color{orange}0.426 & 0.291 \\
    2
    & 0.582 & \color{red}0.615
    & 0.601 & \color{blue}0.618
    & \color{orange}0.614 & 0.584 \\
    3
    & 0.595 & \color{orange}0.631
    & 0.601 & \color{red}0.649
    & 0.563 & \color{blue}0.688 \\
    4
    & 0.694 & 0.644
    & \color{orange}0.718 & 0.695
    & \color{blue}0.763 & \color{red}0.744 \\
    5
    & 0.448 & \color{orange}0.468
    & 0.461 & 0.437
    & \color{red}0.489 & \color{blue}0.495 \\
    6
    & \color{red}0.543 & \color{orange}0.502
    & 0.391 & 0.437
    & 0.441 & \color{blue}0.584 \\
    7
    & \color{blue}0.361 & 0.336
    & \color{red}0.354 & \color{orange}0.339
    & 0.309 & 0.141 \\
    8
    & 0.808 & \color{orange}0.844
    & 0.831 & \color{red}0.855
    & 0.842 & \color{blue}0.860 \\
    9
    & \color{red}0.581 & \color{orange}0.503
    & 0.411 & 0.454
    & \color{blue}0.616 & \color{red}0.581 \\
    10
    & 0.400 & 0.449
    & 0.403 & \color{orange}0.463
    & \color{blue}0.500 & \color{red}0.469 \\
    A
    & 0.865 & 0.885
    & \color{red}0.902 & 0.878
    & \color{orange}0.893 & \color{blue}0.942 \\
    B
    & 0.410 & \color{orange}0.429
    & 0.361 & 0.395
    & \color{blue}0.502 & \color{red}0.439 \\
    C
    & 0.166 & 0.095
    & \color{blue}0.261 & \color{orange}0.127
    & 0.037 & \color{red}0.234 \\
    D
    & 0.699 & \color{orange}0.716
    & 0.685 & 0.695
    & \color{red}0.735 & \color{blue}0.755 \\
    E
    & \color{blue}0.461 & \color{red}0.335
    & 0.328 & 0.313
    & \color{orange}0.330 & 0.018 \\
    F
    & \color{blue}0.598 & \color{red}0.576
    & 0.534 & 0.547
    & 0.541 & \color{orange}0.560 \\
    G
    & 0.974 & \color{blue}0.993
    & 0.976 & 0.981
    & \color{orange}0.989 & \color{red}0.990 \\
    \hline
    $\mu$
    & \color{red}0.556 & \color{red}0.556
    & 0.540 & \color{orange}0.553
    & \color{blue}0.564 & 0.552 \\
    $\mu_w$
    & 0.656 & \color{orange}0.658
    & 0.640 & 0.651
    & \color{red}0.662 & \color{blue}0.690 \\
    \hline
  \end{tabular}

*These results are computed from the published confusion matrix. \\
  
  \vspace{1em}
  \caption{Matthews Correlation Coefficient results for best-performing proposed (Tables~\ref{tab:classwise-top} and~\ref{tab:classwise-no-lm}) and SOTA models (used in Table~\ref{tab:classwise-top}), with and without label merging (LM) for class-wise analysis ({\color{blue}blue}, {\color{red}red}, and {\color{orange}orange} for top 3 results, in descending rank in each category).}
  \label{tab:classwise-mcc}
  \begin{tabular}{c|cc|cc|cc}
    \hline
    & \multicolumn{2}{c|}{FM1}
    & \multicolumn{2}{c|}{FM3}
    & \multicolumn{2}{c}{SOTA} \\
    \hline 
    \multicolumn{7}{c}{17-class scenarios without LM} \\
    \hline
    Model
    & FM1 & FM1\textsubscript{B}
    & FM3 & FM3\textsubscript{B}
    & \textbf{M1}\textsuperscript{*}
    & \textbf{M2} \textsuperscript{*} \\
    $MCC$
    & 0.632 & \color{red}0.652
    & 0.541 & 0.635
    & \color{orange}0.650 & \color{blue} 0.682 \\
    \hline
    \multicolumn{7}{c}{8-class scenarios with LM} \\
    \hline
    Model
    & \multicolumn{2}{c|}{FM1\textsubscript{BL}}
    & \multicolumn{2}{c|}{FM3\textsubscript{BL}}
    & \textbf{M1}\textsuperscript{*}
    & \textbf{M2} \textsuperscript{*} \\
    $MCC$
    & \multicolumn{2}{c|}{\color{red}0.724}
    & \multicolumn{2}{c|}{0.706}
    & \color{orange} 0.719 & \color{blue} 0.750 \\
    \hline    
  \end{tabular}

*These results are computed from the published confusion matrix.
\end{table}

\begin{figure}[tp]
    \centering
    \includegraphics[width=0.7\columnwidth]{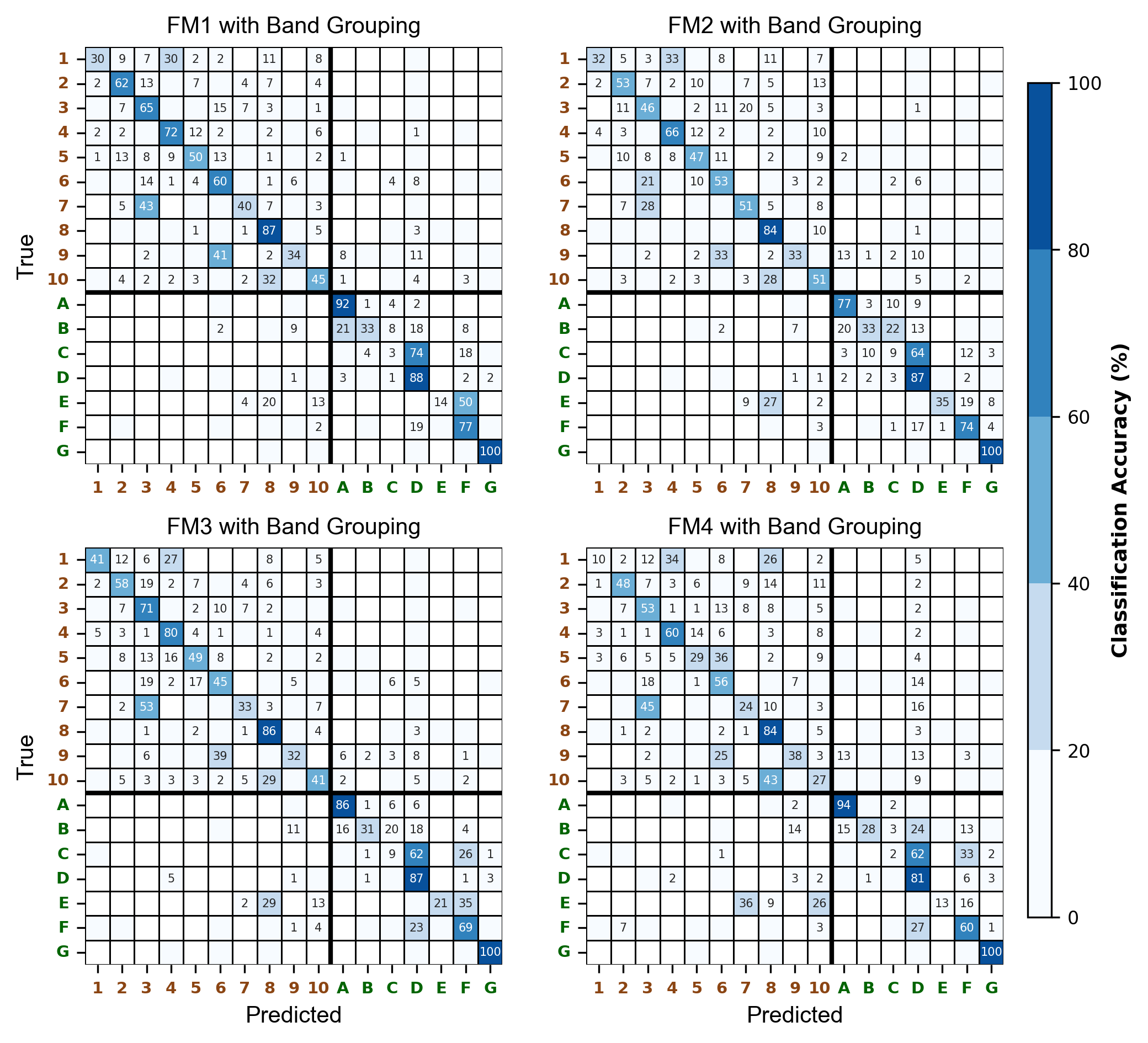}
    \caption{Confusion matrix for LCZ classification outputs of FM1\textsubscript{B}, FM2\textsubscript{B}, FM3\textsubscript{B}, and FM4\textsubscript{B}, \ie~with spectral band grouping.}
    \label{fig:cmat-fm-b}
    \vspace{1em}
    \includegraphics[width=0.7\columnwidth]{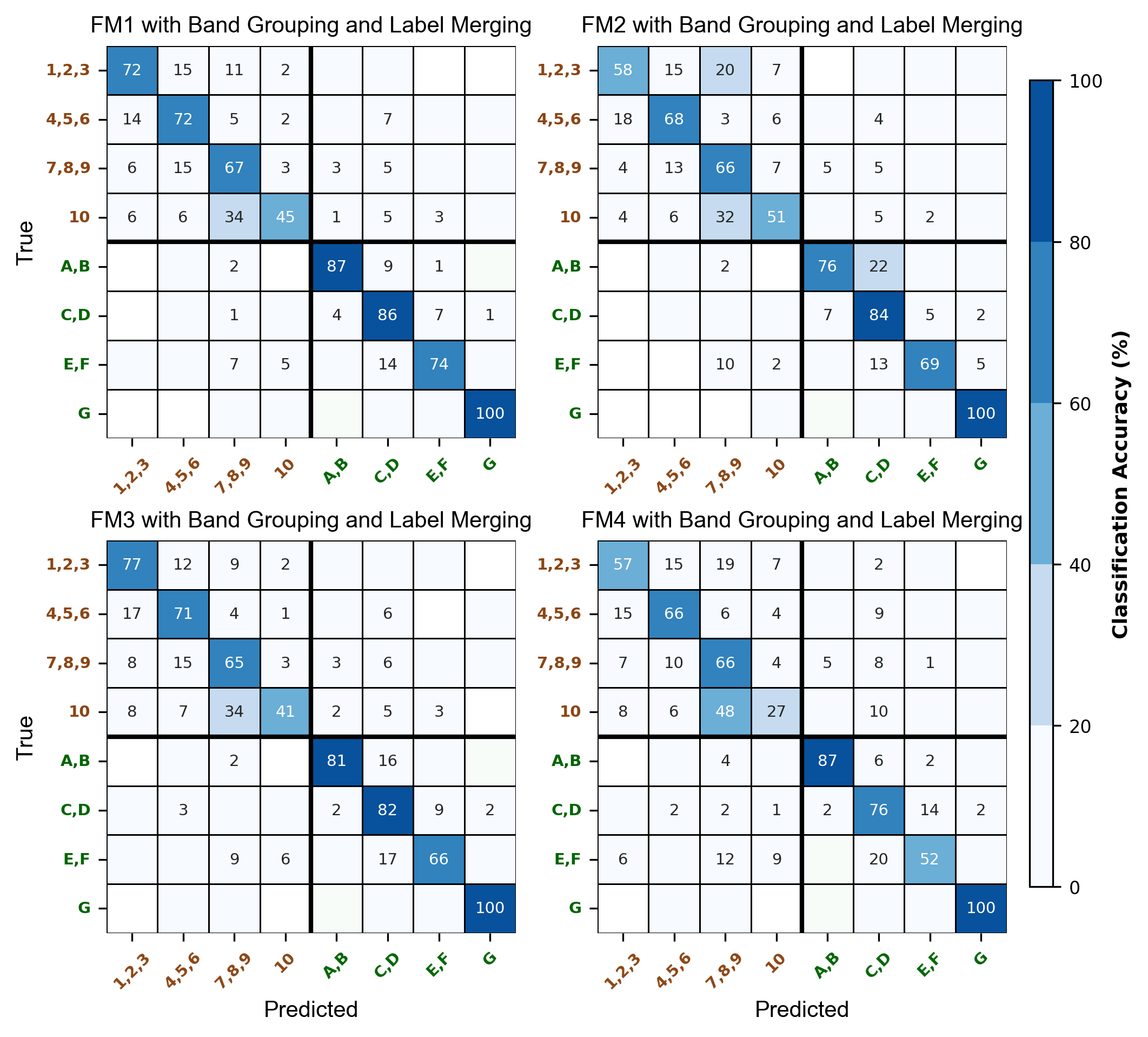}
    \caption{Confusion matrix for LCZ classification outputs for FM1\textsubscript{BL}, FM2\textsubscript{BL}, FM3\textsubscript{BL}, and FM4\textsubscript{BL}, with spectral band grouping and label merging.}
    \label{fig:cmat-fm-bl}
\end{figure}

The confusion matrices in Figure~\ref{fig:cmat-fm-b} show that class C (``Bush, Scrub'') is highly prone to be misclassified as class D (``Low Plants'') in all our models, which is alleviated with label merging, as seen in Figure~\ref{fig:cmat-fm-bl}. We also observe that our models show consistently high accuracies for class 8 (``Large Low-Rise''), class A (``Dense Trees''), and class D (``Water'').

Additional results are given in Appendix~\ref{app:additional-results}. Table~\ref{tab:classwise-additional} shows the performance of 8-class hybrid models, FM1\textsubscript{L}, FM1b\textsubscript{BL}, and FM2b\textsubscript{BL}, which are relatively well-performing, as given in Table~\ref{tab:ablation}. Figures~\ref{fig:cmat-fm} and~\ref{fig:cmat-fm-l} complete the confusion matrices for all models FM* and FM*\textsubscript{L} in addition to Figures~\ref{fig:cmat-fm-b} and~\ref{fig:cmat-fm-bl}, thus, covering all models listed in Table~\ref{tab:overall}.

\clearpage
\section{Conclusion} \label{sec:conclusion}
\begin{figure}[h]
\centering
\includegraphics[width=\columnwidth]{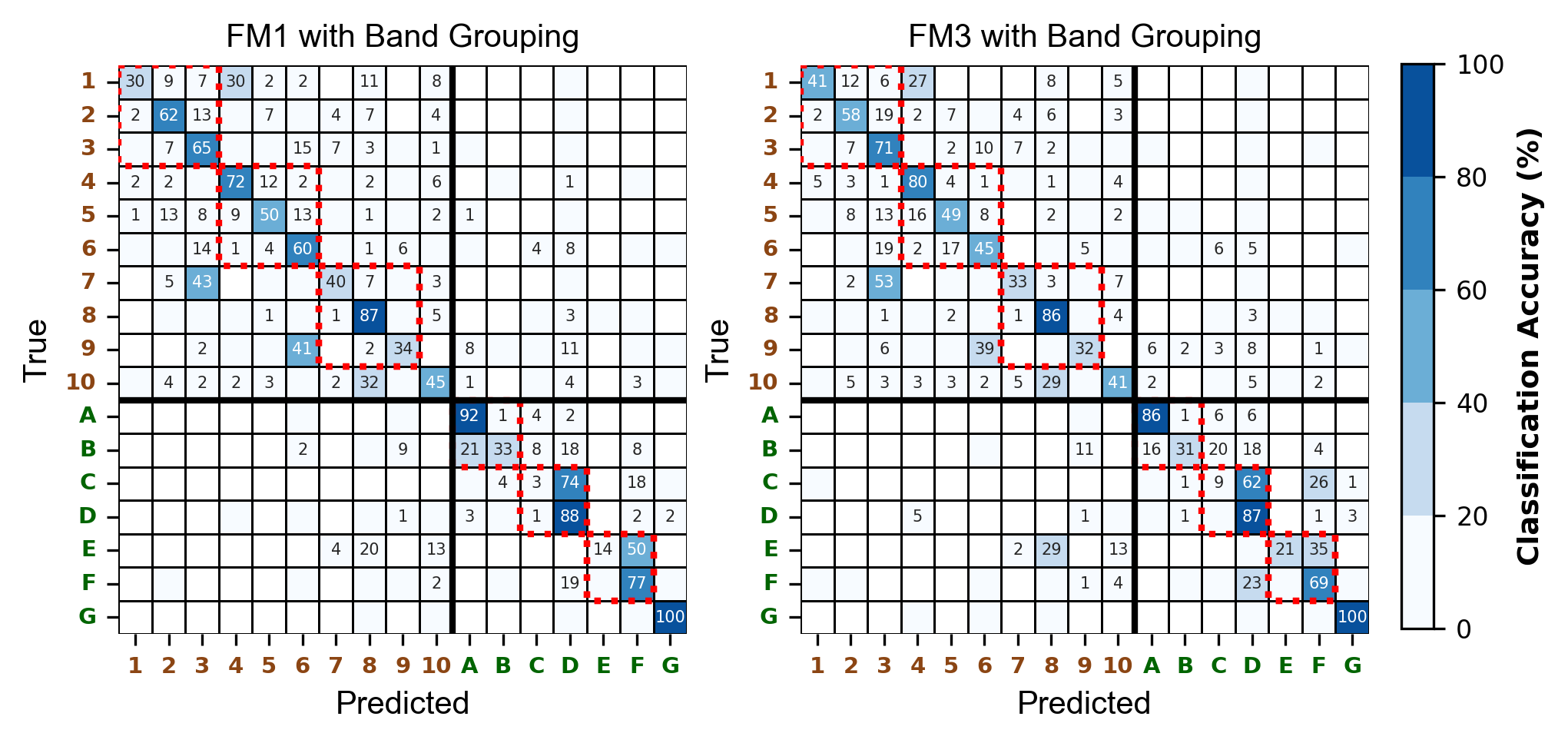}
\caption{The red blocks along the diagonal show how misclassifications get merged as true positives during label merging, thus improving overall accuracy.}
\label{fig:discussion-lm}
\end{figure}

Figure~\ref{fig:discussion-lm} shows how aggregating the submatrices in the confusion matrix converts a 17-class matrix to that of an 8-class matrix. This explains the trivial increase in true positives owing to the merging of the misclassifications shown in the block diagonal.

Our models perform better than the SOTA models, namely, MSCA-MSLCZNet and MSCA-Net, for overall analysis; and outperform MsF-LCZ-Net for class-wise analysis. MSCA-MSLCZNet uses a multiscale coordinate attention-based multistream classification network, and two additional branches of rule-based label corrections for frequently misclassified classes, namely, classes 6-9 and classes C-D. In Table~\ref{tab:classwise-no-lm}, we observe that our models perform better in the macro-average $F_1$ score, as opposed to the weighted average score, compared to the SOTA models. On closer examination, we find that the classes 1, 2, 7, E, and F, which lead to better performance, are relatively underrepresented. Classes 1, 4, 5, 7, B, E, and F have less than 2\% of samples, and class 2 has $\approx$5\% of samples.

The true novelty in our work stems from the use of data-level fusion of SAR and MSI, which differs from using feature maps in both MSCA-Net and MsF-LCZ-Net. The superior performance of MSCA-Net can be attributed to the usage of a single modality, \ie~MSI. Our fusion models are able to match the performance of \textbf{M2}, despite the challenge of using a multimodal dataset, owing to the inclusion of information from the fused data streams in addition to the fused feature maps. The data fusion is able to effectively capture the complementary characteristics of SAR and MSI remote sensing capabilities, such as surface roughness/texture and spectral signature, respectively. This also helps our models, especially with band grouping and hybrid fusion, to effectively classify underrepresented classes.

Summarizing, we demonstrate that multiple levels of early and intermediate fusions followed by a hybrid fusion, along with band grouping and label merging, are comparable to a multistreaming architecture in MSCA-MSLCZNet and a multi-branch CNN in MsF-LCZ-Net. The multiple fusion steps enhance the representation of under-sampled classes, which improves the classifier's performance.

In future work, we intend to examine whether a mixture of experts (MoE) mechanism of hybrid fusion models and multi-branch models can improve all aspects of LCZ classification.

Amongst our proposed models, FM2 and FM4, which use an attention mechanism and late fusion, respectively, are theoretically equipped to perform well in LCZ classification. However, they require special considerations in their implementation to boost their performance. An in-depth study of such improvements is in the scope of future work.

\clearpage
\appendix

  \section{Additional Performance Results}\label{app:additional-results}
Additional results of overall performance metrics of all ablation experiments, namely, Precision ($P$) and Recall ($R$), are given in Table~\ref{tab:additional-results}, thus completing the significant results shown in Table~\ref{tab:ablation} in Section~\ref{sec:results-overall}.

\begin{table}[h]
  \centering
  \caption{Additional ablation experiments supplementing Table~\ref{tab:ablation}.}
  \label{tab:additional-results}
  \begin{tabular}{ccc|cc|ccc|c}
    \hline
    \multicolumn{9}{c}{\bf Overall Accuracy $OA$} \\
    \hline
    FM1 & FM1a & FM1b
    & FM2 & FM2b
    & FM3 & FM3a & FM3b
    & FM4 \\
    0.662 & 0.608 & 0.642
    & 0.660 & 0.623
    & 0.655 & 0.579 & 0.637
    & 0.581 \\
    \hline
    FM1\textsubscript{B} & FM1a\textsubscript{B} & FM1b\textsubscript{B}
    & FM2\textsubscript{B} & FM2b\textsubscript{B}
    & FM3\textsubscript{B} & FM3a\textsubscript{B} & FM3b\textsubscript{B}
    & FM4\textsubscript{B} \\
    0.591 & 0.672 & 0.681 
    & 0.637 & 0.647
    & 0.665 & 0.666 & 0.656
    & 0.624 \\
    \hline
    FM1\textsubscript{L} & FM1a\textsubscript{L} & FM1b\textsubscript{L}
    & FM2\textsubscript{L} & FM2b\textsubscript{L}
    & FM3\textsubscript{L} & FM3a\textsubscript{L} & FM3b\textsubscript{L}
    & FM4\textsubscript{L} \\
    0.737 & 0.699 & 0.726
    & 0.731 & 0.702
    & 0.731 & 0.653 & 0.722
    & 0.664 \\
    \hline
    FM1\textsubscript{BL} & FM1a\textsubscript{BL} & FM1b\textsubscript{BL}
    & FM2\textsubscript{BL} & FM2b\textsubscript{BL}
    & FM3\textsubscript{BL} & FM3a\textsubscript{BL} & FM3b\textsubscript{BL}
    & FM4\textsubscript{BL} \\
    \color{blue}0.766 & 0.670 & \color{red}0.759
    & 0.720 & 0.734
    & \color{orange}0.751 & 0.743 & 0.742
    & 0.699 \\
    \hline

    \hline
    \multicolumn{9}{c}{\bf Precision $P$} \\
    \hline
    FM1 & FM1a & FM1b
    & FM2 & FM2b
    & FM3 & FM3a & FM3b
    & FM4 \\
    0.602 & 0.536 & 0.585       
    & 0.567 & 0.558             
    & 0.592 & 0.511 & 0.571     
    & 0.490 \\                  
    \hline
    FM1\textsubscript{B} & FM1a\textsubscript{B} & FM1b\textsubscript{B}
    & FM2\textsubscript{B} & FM2b\textsubscript{B}
    & FM3\textsubscript{B} & FM3a\textsubscript{B} & FM3b\textsubscript{B}
    & FM4\textsubscript{B} \\
    0.612 & 0.530 & 0.592       
    & 0.576 & 0.579             
    & 0.587 & 0.603 & 0.602     
    & 0.525 \\                  
    \hline
    FM1\textsubscript{L} & FM1a\textsubscript{L} & FM1b\textsubscript{L}
    & FM2\textsubscript{L} & FM2b\textsubscript{L}
    & FM3\textsubscript{L} & FM3a\textsubscript{L} & FM3b\textsubscript{L}
    & FM4\textsubscript{L} \\
    0.708 & 0.675 & 0.697
    & 0.698 & 0.676
    & 0.697 & 0.622 & 0.706
    & 0.643 \\
    \hline
    FM1\textsubscript{BL} & FM1a\textsubscript{BL} & FM1b\textsubscript{BL}
    & FM2\textsubscript{BL} & FM2b\textsubscript{BL}
    & FM3\textsubscript{BL} & FM3a\textsubscript{BL} & FM3b\textsubscript{BL}
    & FM4\textsubscript{BL} \\
    \color{blue}0.740 & 0.624 & \color{orange}0.730
    & 0.693 & 0.713
    & 0.724 & 0.718 & \color{red}0.732
    & 0.643 \\
    \hline

    \hline
    \multicolumn{9}{c}{\bf Recall $R$} \\
    \hline
    FM1 & FM1a & FM1b
    & FM2 & FM2b
    & FM3 & FM3a & FM3b
    & FM4 \\
    0.556 & 0.459 & 0.533
    & 0.554 & 0.527
    & 0.540 & 0.464 & 0.544
    & 0.457 \\
    \hline
    FM1\textsubscript{B} & FM1a\textsubscript{B} & FM1b\textsubscript{B}
    & FM2\textsubscript{B} & FM2b\textsubscript{B}
    & FM3\textsubscript{B} & FM3a\textsubscript{B} & FM3b\textsubscript{B}
    & FM4\textsubscript{B} \\
    0.560 & 0.451 & 0.556
    & 0.549 & 0.540
    & 0.553 & 0.553 & 0.530
    & 0.475 \\
    \hline
    FM1\textsubscript{L} & FM1a\textsubscript{L} & FM1b\textsubscript{L}
    & FM2\textsubscript{L} & FM2b\textsubscript{L}
    & FM3\textsubscript{L} & FM3a\textsubscript{L} & FM3b\textsubscript{L}
    & FM4\textsubscript{L} \\
    \color{orange}0.735 & 0.657 & 0.704
    & 0.725 & 0.713
    & 0.718 & 0.614 & 0.709
    & 0.638 \\
    \hline
    FM1\textsubscript{BL} & FM1a\textsubscript{BL} & FM1b\textsubscript{BL}
    & FM2\textsubscript{BL} & FM2b\textsubscript{BL}
    & FM3\textsubscript{BL} & FM3a\textsubscript{BL} & FM3b\textsubscript{BL}
    & FM4\textsubscript{BL} \\
    \color{blue}0.754 & 0.636 & \color{red}0.741
    & 0.716 & 0.724
    & 0.729 & 0.732 & 0.710
    & 0.663 \\
    \hline
    
    \hline
    \multicolumn{9}{c}{\bf F1 Score $F_1$} \\
    \hline
    FM1 & FM1a & FM1b
    & FM2 & FM2b
    & FM3 & FM3a & FM3b
    & FM4 \\
    0.564 & 0.462 & 0.529
    & 0.546 & 0.512
    & 0.544 & 0.460 & 0.533
    & 0.454 \\
    \hline
    FM1\textsubscript{B} & FM1a\textsubscript{B} & FM1b\textsubscript{B}
    & FM2\textsubscript{B} & FM2b\textsubscript{B}
    & FM3\textsubscript{B} & FM3a\textsubscript{B} & FM3b\textsubscript{B}
    & FM4\textsubscript{B} \\
    0.558 & 0.454 & 0.556
    & 0.543 & 0.535
    & 0.550 & 0.558 & 0.540
    & 0.473 \\
    \hline
    FM1\textsubscript{L} & FM1a\textsubscript{L} & FM1b\textsubscript{L}
    & FM2\textsubscript{L} & FM2b\textsubscript{L}
    & FM3\textsubscript{L} & FM3a\textsubscript{L} & FM3b\textsubscript{L}
    & FM4\textsubscript{L} \\
    0.713 & 0.657 & 0.700
    & 0.707 & 0.688
    & 0.704 & 0.612 & 0.704
    & 0.637 \\
    \hline
    FM1\textsubscript{BL} & FM1a\textsubscript{BL} & FM1b\textsubscript{BL}
    & FM2\textsubscript{BL} & FM2b\textsubscript{BL}
    & FM3\textsubscript{BL} & FM3a\textsubscript{BL} & FM3b\textsubscript{BL}
    & FM4\textsubscript{BL} \\
    \color{blue}0.744 & 0.626 & \color{red}0.733
    & 0.700 & 0.717
    & \color{orange}0.723 & 0.722 & 0.716
    & 0.650 \\
    \hline

    \hline
    \multicolumn{9}{c}{\bf Kappa Coefficient $\kappa$} \\
    \hline
    FM1 & FM1a & FM1b
    & FM2 & FM2b
    & FM3 & FM3a & FM3b
    & FM4 \\
    0.630 & 0.580 & 0.609
    & 0.628 & 0.588
    & 0.620 & 0.538 & 0.603
    & 0.540 \\
    \hline
    FM1\textsubscript{B} & FM1a\textsubscript{B} & FM1b\textsubscript{B}
    & FM2\textsubscript{B} & FM2b\textsubscript{B}
    & FM3\textsubscript{B} & FM3a\textsubscript{B} & FM3b\textsubscript{B}
    & FM4\textsubscript{B} \\
    0.650 & 0.550 & 0.640
    & 0.603 & 0.614
    & 0.633 & 0.633 & 0.622
    & 0.587 \\
    \hline
    FM1\textsubscript{L} & FM1a\textsubscript{L} & FM1b\textsubscript{L}
    & FM2\textsubscript{L} & FM2b\textsubscript{L}
    & FM3\textsubscript{L} & FM3a\textsubscript{L} & FM3b\textsubscript{L}
    & FM4\textsubscript{L} \\
    0.689 & 0.643 & 0.675
    & 0.682 & 0.648
    & 0.682 & 0.588 & 0.670
    & 0.644 \\
    \hline
    FM1\textsubscript{BL} & FM1a\textsubscript{BL} & FM1b\textsubscript{BL}
    & FM2\textsubscript{BL} & FM2b\textsubscript{BL}
    & FM3\textsubscript{BL} & FM3a\textsubscript{BL} & FM3b\textsubscript{BL}
    & FM4\textsubscript{BL} \\
    \color{blue}0.723 & 0.611 & \color{red}0.715
    & 0.669 & 0.685
    & \color{orange}0.705 & 0.695 & 0.694
    & 0.644 \\
    \hline

    \end{tabular}
\end{table}

\begin{figure}[tp]
    \centering
    \includegraphics[width=0.7\columnwidth]{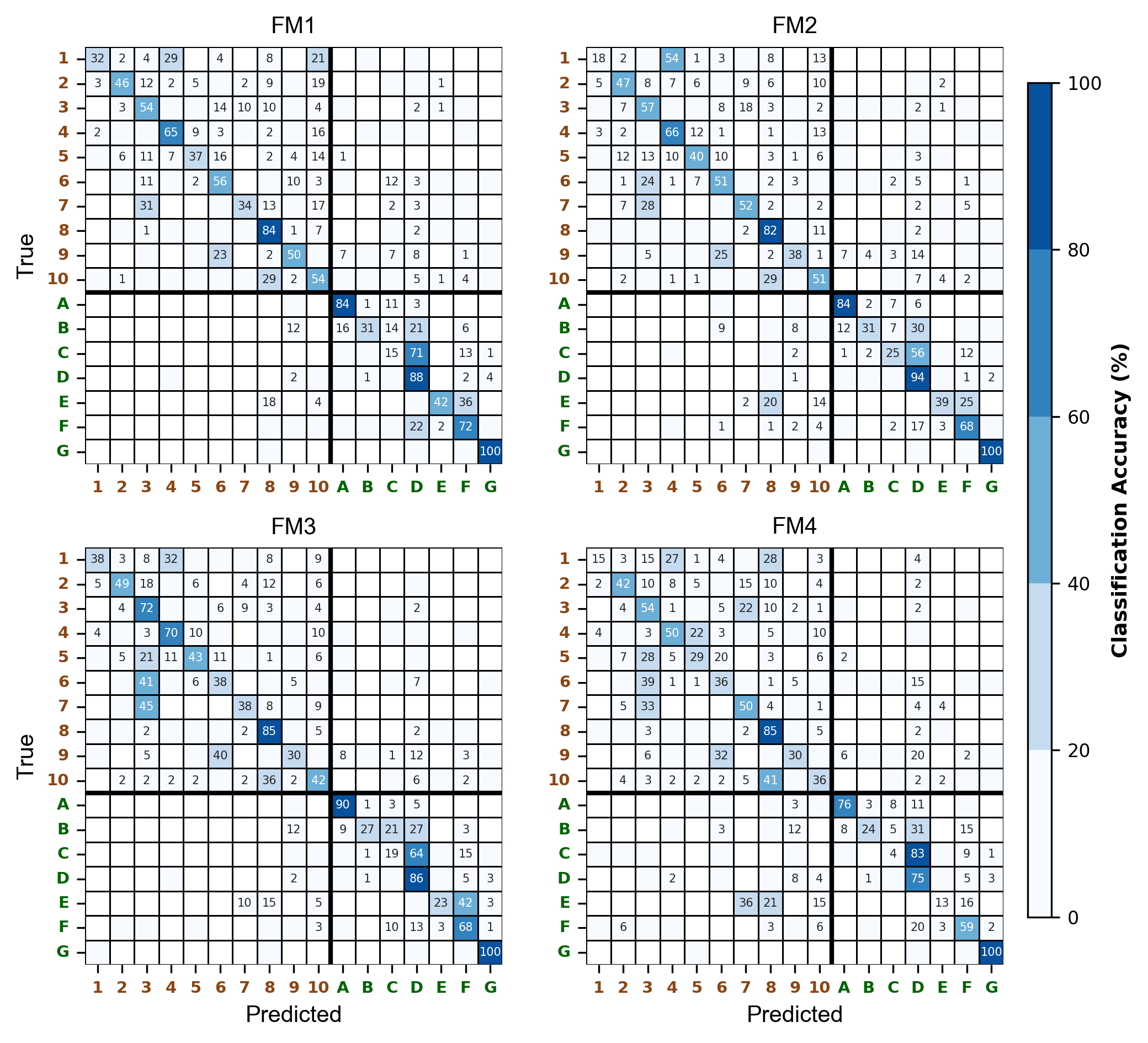}
    \caption{Confusion matrix for LCZ classification outputs of our proposed models, namely, FM1, FM2, FM3, and FM4.}
    \label{fig:cmat-fm}
    \centering
    \vspace{1em}
    \includegraphics[width=0.7\columnwidth]{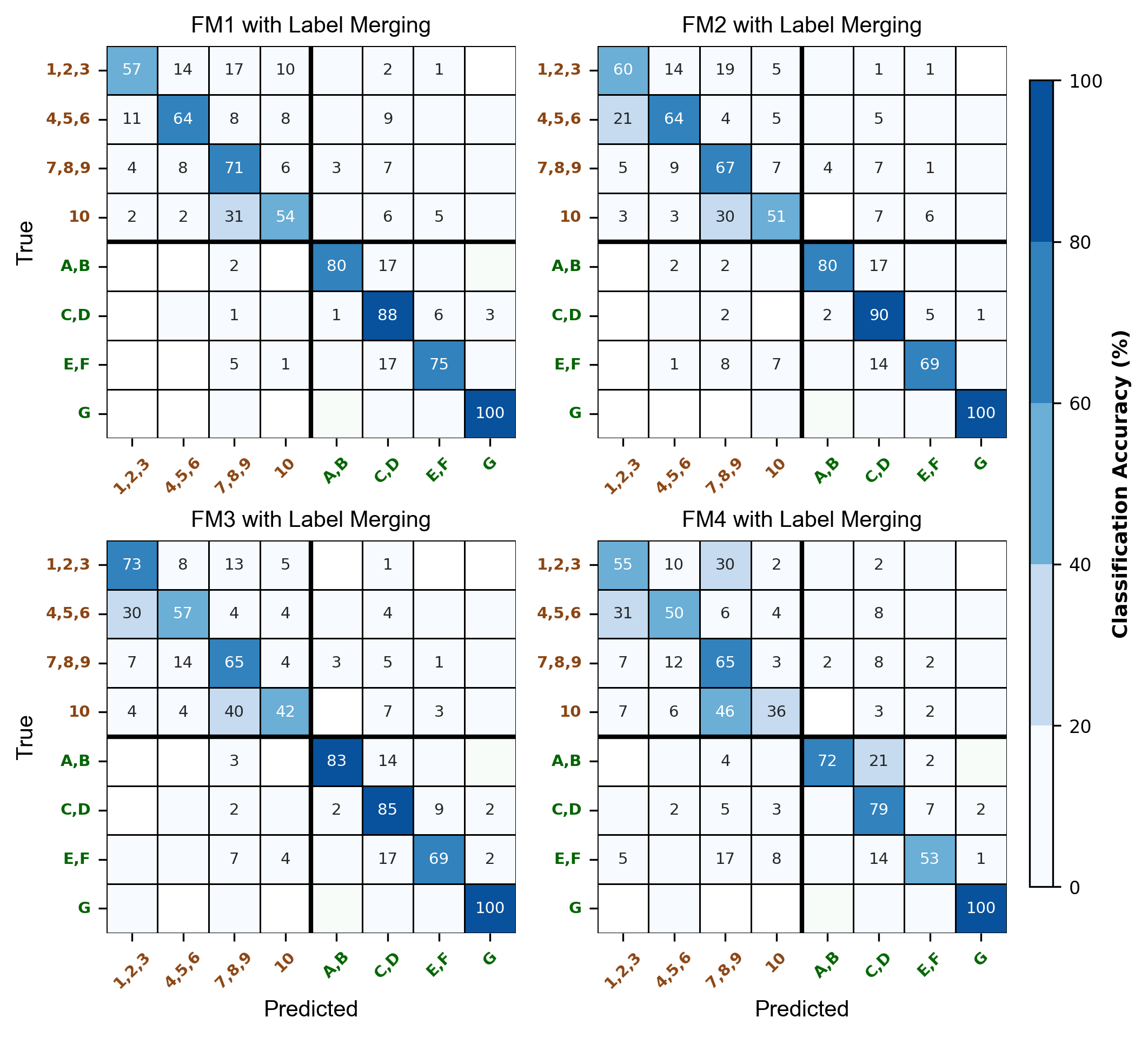}
    \caption{Confusion matrix for LCZ classification outputs for our models with label merging, \ie~FM1\textsubscript{L}, FM2\textsubscript{L}, FM3\textsubscript{L}, and FM4\textsubscript{L}.}
    \label{fig:cmat-fm-l}
\end{figure}

\clearpage
Additional results of class-wise performance metrics of the well-performing fusion models, FM1\textsubscript{L}, FM1b\textsubscript{BL}, and FM2b\textsubscript{BL} (from Tables~\ref{tab:overall} and~\ref{tab:ablation}), with the grouping strategies, are given in Table~\ref{tab:classwise-additional}. Figures~\ref{fig:cmat-fm},~\ref{fig:cmat-fm-b}, and~\ref{fig:cmat-fm-l} provide confusion matrices for the hybrid models, the same with spectral band grouping, and the same with label merging, respectively. This completes the significant results given in Section~\ref{sec:results-classwise}.

\begin{table}[htp]
  \centering
  \caption{Performance results of our best-performing models (from Table~\ref{tab:ablation}), along with their averages, \ie~macro ($\mu$) and weighted ($\mu_w$) ({\color{blue}blue} is the best performance value). Since all the top models use LM, the labels show the original classes used for merging.}
  \label{tab:classwise-additional}
  \begin{tabular}{c|ccc|ccc}
    \hline
    Label
    & FM1\textsubscript{L}
    & FM1b\textsubscript{BL}
    & FM2b\textsubscript{BL}
    & FM1\textsubscript{L}
    & FM1b\textsubscript{BL}
    & FM2b\textsubscript{BL} \\
    \hline
    & \multicolumn{3}{c|}{F1 Score $F_1$}
    & \multicolumn{3}{c}{Kappa Coefficient $\kappa$} \\
    \hline
    \{1,2,3\}
    & 0.663 & \color{blue}0.732 & 0.632 
    & 0.609 & \color{blue}0.680 & 0.564 
    \\
    \{4,5,6\}
    & 0.655 & \color{blue}0.683 & 0.642 
    & 0.598 & \color{blue}0.627 & 0.584 
    \\
    \{7,8,9\}
    & \color{blue}0.734 & 0.720 & 0.718 
    & \color{blue}0.650 & 0.636 & 0.625 
    \\
    \{10\}
    & 0.400 & 0.449 & \color{blue}0.500 
    & 0.370 & 0.430 & \color{blue}0.480 
    \\
    \{A,B\}
    & 0.846 & \color{blue}0.871 & 0.821 
    & 0.827 & \color{blue}0.854 & 0.834 
    \\
    \{C,D\}
    & 0.767 & 0.772 & \color{blue}0.781 
    & 0.720 & \color{blue}0.728 & 0.716 
    \\
    \{E,F\}
    & \color{blue}0.662 & 0.642 & 0.648 
    & \color{blue}0.650 & 0.629 & 0.636 
    \\
    \{G\}
    & \color{blue}0.998 & 0.993 & 0.993 
    & 0.971 & \color{blue}0.992 & \color{blue}0.992 
    \\
    $\mu$
    & 0.713 & \color{blue}0.733 & 0.717 
    & 0.674 & \color{blue}0.697 & 0.677 
    \\
    $\mu_w$
    & 0.739 & \color{blue}0.758 & 0.732 
    & 0.690 & 0.710 & 0.679 
    \\
    \hline
    \hline
    & \multicolumn{3}{c|}{Precision $P$}
    & \multicolumn{3}{c}{Recall $R$} \\
    \hline
    \{1,2,3\}
    & \color{blue}0.785 & 0.746 & 0.669 
    & 0.574 & \color{blue}0.719 & 0.600 
    \\
    \{4,5,6\}
    & \color{blue}0.677 & 0.666 & \color{blue}0.677 
    & 0.635 & 0.701 & \color{blue}0.710 
    \\
    \{7,8,9\}
    & 0.758 & \color{blue}0.774 & 0.741 
    & \color{blue}0.712 & 0.673 & 0.691 
    \\
    \{10\}
    & 0.317 & \color{blue}0.506 & 0.484 
    & \color{blue}0.540 & 0.403 & 0.518 
    \\
    \{A,B\}
    & \color{blue}0.901 & 0.857 & 0.815 
    & 0.797 & \color{blue}0.885 & 0.827 
    \\
    \{C,D\}
    & 0.682 & \color{blue}0.714 & 0.695 
    & 0.875 & 0.840 & \color{blue}0.890 
    \\
    \{E,F\}
    & 0.594 & 0.585 & \color{blue}0.635 
    & \color{blue}0.749 & 0.712 & 0.662 
    \\
    \{G\}
    & 0.951 & \color{blue}0.990 & 0.989 
    & \color{blue}0.998 & 0.996 & 0.996 
    \\
    $\mu$
    & 0.708 & \color{blue}0.730 & 0.713 
    & 0.735 & \color{blue}0.741 & 0.724 
    \\
    $\mu_w$
    & 0.754 & \color{blue}0.761 & 0.734 
    & 0.740 & \color{blue}0.759 & 0.734 
    \\
    \hline
  \end{tabular}
\end{table}

\clearpage
\bibliographystyle{unsrt}
\bibliography{papers_sar-msi-fusion}
\end{document}